\documentclass[journal]{IEEEtran}

\usepackage{amsmath,amsfonts}
\usepackage{algorithmic}
\usepackage{algorithm}
\usepackage{array}
\usepackage{subcaption}
\usepackage{textcomp}
\usepackage{stfloats}
\usepackage{url}
\usepackage{verbatim}
\usepackage{graphicx}
\usepackage{cite}

\usepackage{xcolor}
\usepackage{colortbl}
\usepackage{flushend}
\usepackage[utf8]{inputenc}
\usepackage[draft]{hyperref}
\usepackage{booktabs}

\DeclareGraphicsExtensions{.pdf,.svg,.jpg,.png,.eps} 
\graphicspath{{./assets/}} 

\def\BibTeX{{\rm B\kern-.05em{\sc i\kern-.025em b}\kern-.08em
    T\kern-.1667em\lower.7ex\hbox{E}\kern-.125emX}}

\begin{document}

\title{Self-Predictive Representation for\\Autonomous UAV Object-Goal Navigation}


\author{Angel Ayala$^{1}$,
Donling Sui$^{3}$,
Francisco Cruz$^{2,4}$,
Mitchell Torok$^{3}$,\\
Mohammad Deghat$^{3}$,
and Bruno J. T. Fernandes$^{1}$
\thanks{$^{1}$Angel Ayala and Bruno J. T. Fernandes are with the Escola Polit\'ecnica de Pernambuco, Universidade de Pernambuco, Recife, Brasil {\tt\small {aaam, bjtf}@ecomp.poli.br}}%
\thanks{$^{2,4}$Francisco Cruz is with the School of Computer Science and Engineering, University of New South Wales, Sydney, Australia and Escuela de Ingenier\'ia, Universidad Central de Chile, Santiago, Chile {\tt\small f.cruz@unsw.edu.au}}
\thanks{$^{3}$Donling Sui, Mitchell Torok, and Mohammad Deghat are with the School of Mechanical and Manufacturing Engineering, University of New South Wales, Sydney, Australia}
}

\markboth{Submitted to IEEE TRANSACTIONS ON ROBOTICS}{Ayala et al. SPR for autonomus UAV navigation}


\maketitle

\begin{abstract}
Autonomous Unmanned Aerial Vehicles (UAVs) have revolutionized industries through their versatility with applications including aerial surveillance, search and rescue, agriculture, and delivery.
Their autonomous capabilities offer unique advantages, such as operating in large open space environments.
Reinforcement Learning (RL) empowers UAVs to learn intricate navigation policies, enabling them to optimize flight behavior autonomously.
However, one of its main challenge is the inefficiency in using data sample to achieve a good policy.
In object-goal navigation (OGN) settings, target recognition arises as an extra challenge.
Most UAV-related approaches use relative or absolute coordinates to move from an initial position to a predefined location, rather than to find the target directly.
This study addresses the data sample efficiency issue in solving a 3D OGN problem, in addition to, the formalization of the unknown target location setting as a Markov decision process.
Experiments are conducted to analyze the interplay of different state representation learning (SRL) methods for perception with a model-free RL algorithm for planning in an autonomous navigation system.
The main contribution of this study is the development of the perception module, featuring a novel self-predictive model named AmelPred.
Empirical results demonstrate that its stochastic version, AmelPredSto, is the best-performing SRL model when combined with actor-critic RL algorithms.
The obtained results show substantial improvement in RL algorithms' efficiency by using AmelPredSto in solving the OGN problem.
\end{abstract}


\section{Introduction}

Autonomous Unmanned Aerial Vehicles (UAVs) are used extensively in tasks such as aerial surveillance, search and rescue missions, agricultural monitoring, and package delivery~\cite{liu2020unmanned,mohsan2022towards,del2021unmanned}.
The UAV control problem, has reregularly been addressed as a two-level controller:
1) the inner-loop controller, which deals with vehicle attitude stabilization,
and 2) the outer-loop controller, which deals with the UAV displacement~\cite{sonugur2023review}.
For attitude control, it is a common practice to use autopilot boards such as Ardupilot or Pixhawk~\cite{perez2021introducing}, or implement a simple PID controller~\cite{hua2013introduction}.
The outer-loop controller, in contrast, is responsible for high-level decision making, by processing the current vehicle's state and raw surrounding information to choose where to move next~\cite{rao2022position,cardenas2023intelligent}.
As raw sensor data could present incorrect measurements, more advanced sensor processing approaches are required for control~\cite{bijjahalli2020advances}.

Suitable autonomous outer-loop control methods are challenging when there is no accurate model for vehicle dynamics due to its non-linearity, in addition to presence of different objects in the environment.
Reinforcement Learning (RL) algorithms have successful been proven in autonomous drone control by addressing the problem as Markov Decision Processes (MDP)~\cite{sutton2018reinforcement}.
Many RL studies use a model-based approach~\cite{azar2021drone} or a reduced action space~\cite{fu2022memory} to fulfill the navigation problem without collision satisfactorily.
Other studies utilize deep RL for visual-based observations, which involves a higher complexity level due to the raw-image data dimensionality~\cite{lu2018survey}.
However, RL faces open challenges in both MDP and Partially Observed MDP (POMDP), where the agent cannot directly measure some information, heading to sample-data inefficiencies by requiring too much time to achieve a suitable policy.

Moreover, the object-goal navigation (OGN) problem requires displacing a robot over a particular area to search for a target object~\cite{ayala2024uav,du2023object}.
Most OGN studies have focused on embodied indoor systems~\cite{kotar2023entl}, aiming to achieve a spatial representation from structured scenarios~\cite{zhao2020learning,dugas2021navrep}.
The earliest studies demonstrated that informative features increase performance in solving OGN problems~\cite{llofriu2015goal}.
Current literature proposes decoupling representation from policy learning to improve sample efficiency and learning time~\cite{lesort2018state}.
In this regard, representations for OGN must portray the internal vehicle state and the target-related information in a lower-dimensional latent vector~\cite{sun2024survey}.

Representation learning is a machine learning process that creates a compact and informative vector in a lower dimension~\cite{bengio2013representation}.
State Representation Learning (SRL) then compresses the observation into a state vector in such a way that it attends to the Markov property~\cite{allen2021learning}.
One of the first representation methods to enhance RL performance in navigation, named as Relative Line Position Representation (RLPR), transforms laser sensor readings into the distance of different object classes related to color and neighborhood~\cite{frommberger2007generalizing}.
Later methods, which address transformation over the latent vector obtained from a Variational AutoEncoder~\cite{jonschkowski2015learning}, known as robotic priors, have proven useful, particularly in unique position target contexts~\cite{lesort2018state,bijman2020state}.
In current methods, the representation stage is performed by an encoder function that compresses the sensor readings into a latent representation, commonly designed for visual-based~\cite{hoeller2021learning,xue2023monocular,huang2023representation} or multi-modal approaches~\cite{huang2019transferable,bonatti2020learning,vemprala2021representation}.

Previous studies have already demonstrated the effectiveness of RL algorithms combined with SRL on classical RL benchmarks, such as Atari and DMControl~\cite{tang2023understanding,liu2024enhancing}.
This study may be the first to propose an RL-SRL combination for solving OGN problems using an aerial robotics platform, such as a quadcopter.
In this regard, this paper first introduces the background of RL algorithms and SRL methods in Section~\ref{section:background}.
Next, our novel SRL architecture, based on the self-predictive approach and experimental settings, is presented in Section~\ref{section:proposal}.
Subsequently, Section~\ref{section:environment} details the benchmark environment, followed by the corresponding results report and discussion in Section~\ref{section:results}.
Finally, we summarize our findings and discuss future work in Section~\ref{section:conclusion}.
The overall contributions of this study can be summarized as:

\begin{itemize}
    \item A comprehensive study about the effect on self-predictive representations in RL algorithms for solving an object-goal autonomous problem with a quadcopter.
    \item A novel continuous RL agnostic SRL method for sample-efficient autonomous navigation of a quadrotor system.
    \item A fully dimensional quadcopter control problem modeled by an intuitive reward function, under an object-goal navigation context.
    \item A publicly available 3D simulated benchmark for UAV object-goal autonomous navigation on Webots.
\end{itemize}

\section{Background and related works}~\label{section:background}

Reinforcement learning~\cite{sutton2018reinforcement} refers to animal behavior psychology, which defines learning as an iterative trial and error-process.
A goal-seeking agent must discover a suitable action policy to solve a given problem at each iteration.
Complementarily, state representation learning decouples the representation and policy optimization from the RL framework.
Representation learning is an extraction process of useful information from raw data input, by enhancing the latent representation from a noisy data source at a lower dimension~\cite{bengio2013representation}.
Hence, the latent representation should be compact and, at the same time, informative.
Therefore, SRL focuses on learning low dimensional features, which evolve through time by the agent's actions influence~\cite{lesort2018state}, improving data-sample efficiency.

\subsection{Reinforcement learning algorithms}
The RL framework, as a machine learning approach, can efficiently solve Markov decision processes (MDP), where an agent in a given state $s_t \in S$ can perform an action $a_t \in A$, transiting to a new state $s_{t+1} \in S$, and obtaining a reward value $r_{t+1} \in R$, after which the process repeats.
The states, actions, and rewards ($S, A, R$) can be in a discrete or continuous domain.
The process of selecting an action is known as policy, which has been proven to converge into an optimal behavior, through Temporal-difference (TD) algorithm.
TD(0), the most simple version of TD, enhances the policy by iterating over the approximation function, known as a policy evaluation or prediction problem.
A well-known TD algorithm is $Q$-learning~\cite{watkins1992q}, which learns to approximate an action-value function in an off-policy way, formalized by:
\begin{equation}\label{eq:Qlearning}
  Q(s_t, a_t) \gets Q(s_t, a_t) + \alpha \big[r_{t+1} + \gamma \underset{a}{\max}Q(s_{t+1}, a) - Q(s_t, a_t)\big],
\end{equation}
where $\alpha$ is the learning rate, and $\gamma$ is a discount factor determining the future rewards importance.
These Q-values indicate how much reward will be obtained in the future by choosing that action.

Most real-world problems are too complex to learn as finite state-action pairs, so they are approached with parameterized action-value functions instead~\cite{mnih2015human}.
The first selected algorithm for comparison was deep $Q$-network (DQN), a multi-layered neural network that outputs action values $Q(s_t, \cdot ;\theta_t)$ for a given state $s_t$ with $\theta$ parameters at step $t$.
The neural network then maps an $n$-dimensional state space to $m$ actions by updating $Q(s_t,a_t;\theta_t)$ towards a TD target value $\mathrm{y}_{t}^{Q}$ using $Q$-learning, formalized as:
\begin{equation}\label{eq:policyOptimization}
    \text{J}_\pi(Q) =
      \mathbb{E}_{s_t, a_t \sim \pi}\big[ \mathrm{y}_{t}^{\text{DQN}} - Q(s_t, a_t; \theta_t) \big],
\end{equation}
where $\mathrm{y}_{t}^{\text{DQN}} = r_{t+1} + \gamma \underset{a}{\max}Q(s_{t+1}, a;\theta'_t)$, with $\theta'_t$ being the target network parameters.

Most recent algorithms implements actor-critic architecture by first decoupling policy and value functions learning.
Hence, the actor model learns a policy for the action selection sequence, defining the agent's behavior.
In contrast, the critic model learns to estimate a value function, describing how much future reward can be obtained next.
Another algorithm used in this study was the Twin-Delayed DDPG (TD3)~\cite{fujimoto2018addressing} algorithm, a deterministic approach which learns two $Q$-value functions through mean square Bellman error minimization, defining the critic loss for non-terminal cases as:
\begin{equation}\label{eq:td3_critic_target}
    \begin{array}{rcl}
J_{Q_i} & = & \underset{(s,a,r,s')\sim\mathcal{D}}{\mathbb{E}}
    \big[(Q_{\theta_i}(s,a) - y^{\text{TD3}})^2 \big] \\
y^{\text{TD3}} & = & r + \gamma \min_{i=1,2} Q_{\phi'_{i}}(s', a'(s'))
\end{array},
\end{equation}
with $a'(s')$ being a clipped noisy target next action value.
Additionally, the policy loss function is then defined by:
\begin{equation}\label{eq:td3_policy}
    J_\pi = \underset{s\sim\mathcal{D}}{\mathbb{E}}
    \big[Q_{\theta_1}(s,\pi(s)) \big].
\end{equation}
The last algorithm used was the Soft Actor-Critic (SAC)~\cite{haarnoja2018soft}, which optimizes a stochastic policy by maximizing a trade-off between reward and policy entropy by defining the loss function as:
\begin{equation}\label{eq:sac_critic}
    \begin{array}{rcl}
J_Q & = & \underset{(s,a,r,s')\sim\mathcal{D}}{\mathbb{E}}
    \big[\frac{1}{2}\sum^{2}_{i=1}(Q_{\theta_i}(s,a) - y^{\text{SAC}})^2 \big] \\
y^{\text{SAC}} & = & r + \gamma \left( \min_{j=1,2} Q_{\phi_{\text{targ},j}}(s', \tilde{a}') - \alpha \log \pi_{\theta}(\tilde{a}'|s') \right)
\end{array},
\end{equation}
with $\tilde{a}' \sim \pi_{\theta}(\cdot|s')$ being the next stochastic action.
Additionally, the policy loss function is then defined by:
\begin{equation}\label{eq:sac_policy}
    J_\pi = \underset{s\sim\mathcal{D}}{\mathbb{E}}
    \big[\min_{j=1,2}Q_{\theta_j}(s,\tilde{a}) - \alpha \log \pi(\tilde{a}|s) \big].
\end{equation}

\subsection{Self-predictive representations}
Self-predictive representation approaches arise from self-supervised learning, which optimizes an encoder model through a similarity measure between a given sample and a positive and negative subset~\cite{ericsson2022self}.
The core idea behind the optimization process is to bring the sample closer to the positive subset while keeping it away from the negative subset, as seen in contrastive methods~\cite{oord2018representation,tsai2021self}.
For example, the BYOL~\cite{grill2020bootstrap} architecture comprises two neural networks, referred to as online and target networks, which learn from each other through interaction.
The online network grabs an image sample to predict the target network representation from the sample's augmented version, as the positive subset.
A study has already implemented this architecture with RL, known as Augmented Temporal Contrast (ATC)~\cite{stooke2021decoupling}, where the online model processes the current observation and is updated using backpropagation.
The target model is applied to the following five observations and updated by a moving average method known as stop-gradient.

Recently, a self-predictive representations (SPR) method has been successfully applied in solving Atari-based environments using the Rainbow algorithm~\cite{schwarzer2020data}.
Similar to BYOL and ATC, SPR processes the input using encoder and projection models from both the online and target networks.
In contrast, SPR adds an extra model, named the transition model, which receives the current latent vector and action in the online network, located between the encoder and the projection models.
The central idea of SPR is to learn a representation in the latent space by minimizing the prediction error.
However, error minimization does not ensure a good representation since it suffers from representational collapse, delivering trivial information due to a lack of the true ground-truth~\cite{tang2023understanding,ni2024bridging}.
In the study by Ni et al.~\cite{ni2024bridging}, it is proven both theoretically and empirically that stop-gradient prevents representational collapse and proposes a minimalist implementation.
Nevertheless, the experimental setting used was MuJoCo continuous control problems, and its implementation was based solely on the TD3 algorithm.
In this regard, this study proposes a wider analysis of the performance impact of self-predictive approaches in object-goal navigation problems.

\subsection{State representation for quadcopter control}\label{section:SRL_navigation}
The study of LatentSLAM~\cite{ccatal2021latentslam} can be considered an approach of SPR for Simultaneous Localization and Mapping (SLAM), where a generative prior and posterior jointly learn the underlying latent vector in a bootstrapped fashion.
Studies related to quadcopter navigation that already use SRL are focused on visual-based observations~\cite{zhao2023learning,xue2023monocular,yue2024semantic}, using a staged approach with a pretrained AutoEncoder model from computer vision datasets or expert demonstrations.
However, those studies suffer from the limitations common to deep learning methods, such as limited generalization capabilities conditioned beyond the training data, which can be expensive to collect or may contain wrongly-labeled samples.
Additionally, data driven approaches can introduce bias from the object-of-interest features in the training subset, affecting the quality of downstream tasks, such as decision-making.
As current SRL literature suggests the use of an SPR approach to enhance sample efficiency in RL algorithms, this study aims to implement a novel SPR architecture to efficiently solve an OGN problem.

\section{SRL for autonomous navigation}~\label{section:proposal}
\begin{figure}
    \centering
    \includegraphics[width=0.9\linewidth]{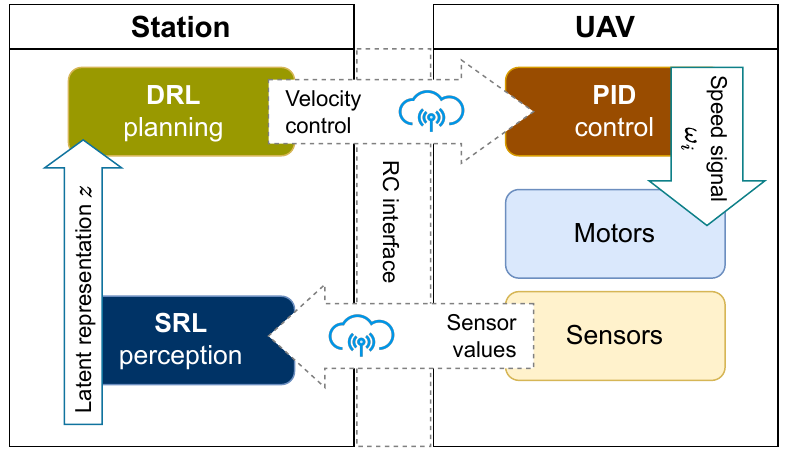}
    \caption[The overall architecture of Chemamuy ANS.]{The overall architecture of Chemamuy ANS for solving control, perception, and planning tasks with a quadrotor.
    On the UAV side, a PID-based control system deals with the UAV dynamics while continuously transmitting its sensor readings.
    On the Station side, the sensor data is processed by an SRL-based perception method, creating a latent representation $z$ further processed by RL.
    The RL approach sends the desired velocities to displace the UAV towards the target.}
    \label{fig:overallArchitecture}
\end{figure}

The proposed SRL method was projected as a perception module inside a mapless object-goal Autonomous Navigation System (ANS) named Chemamuy~\footnote{The name ``Chemamuy'' was inspired by the Mapuche people, and means ``someone is moving towards that.''
Chemamuy was derived from the Mapudungun grammar published at https://mapudungun.cl/categories/gramatica.html and processed by an LLM, for searching and processing possible words, resulting in one such meaning that emphasizes intentional movement, autonomy, and adaptive decision-making of an agent towards a target object.}.
The main purpose of the learning-based perception module is to enhance quadcopter navigation performance by delivering a useful latent information.
The overall Chemamuy's architecture is depicted in Figure~\ref{fig:overallArchitecture}, comprising two sides.
On one side, the UAV handles the control task; on the other, the station handles the planning and perception tasks.
A radio communication (RC) interface transports the action signal and sensor values between the station and the UAV.
In this regard, Chemamuy's architecture complies with the Increasing Precision Decreasing Intelligence principle~\cite{saridis1989analytic}, where a PID-based method addresses the vehicle velocity control belonging to the execution level.
For the coordination level, the core of our proposal, the Chemamuy's SRL-based perception method asynchronously processes the sensor data to create a compressed representation $z$ about the current vehicle state and surroundings.
Finally, at the organizational level, Chemamuy's RL-based path planning method processes the $z$ vector into the desired velocity control signal and transmits it back to the UAV.
In this regard, Chemamuy's performance considers two main constraints: a limited flight area due to the communication range with the base station and a the real-time control processing for safety assurance.

\subsection{Self-predictive representation}

The proposed self-predictive approach, named AmelPred\footnote{The name ``AmelPred'' arises from the Mapudungun word ``Amelkantun'' that means ``to represent something'' and the English word ``Prediction''.}, was inspired by the SPR architecture from \cite{schwarzer2020data}.
Major differences between SPR and AmelPred are related to the addressed observation space and the problem solved.
SPR was projected to process a visual-based observation for Atari-based problems.
In contrast, AmelPred processess a vector-based observation for solving an OGN problem.
Recent literature also suggests the use of stochastic functions given the uncertainty present in sensor processing and historical experiences~\cite {ni2024bridging}.
In this regard, a deterministic and stochastic encoder definitions of AmelPred, denoted as AmelPredDet and AmelPredSto, are included.
The AmelPred architecture aims to be lightweight and efficient for different RL algorithms, unlike \cite{schwarzer2020data} and \cite{ni2024bridging} studies, which only addresses Rainbow-based and TD3-based experiments, respectively.

\begin{figure*}
    \centering
    \includegraphics[width=0.9\linewidth]{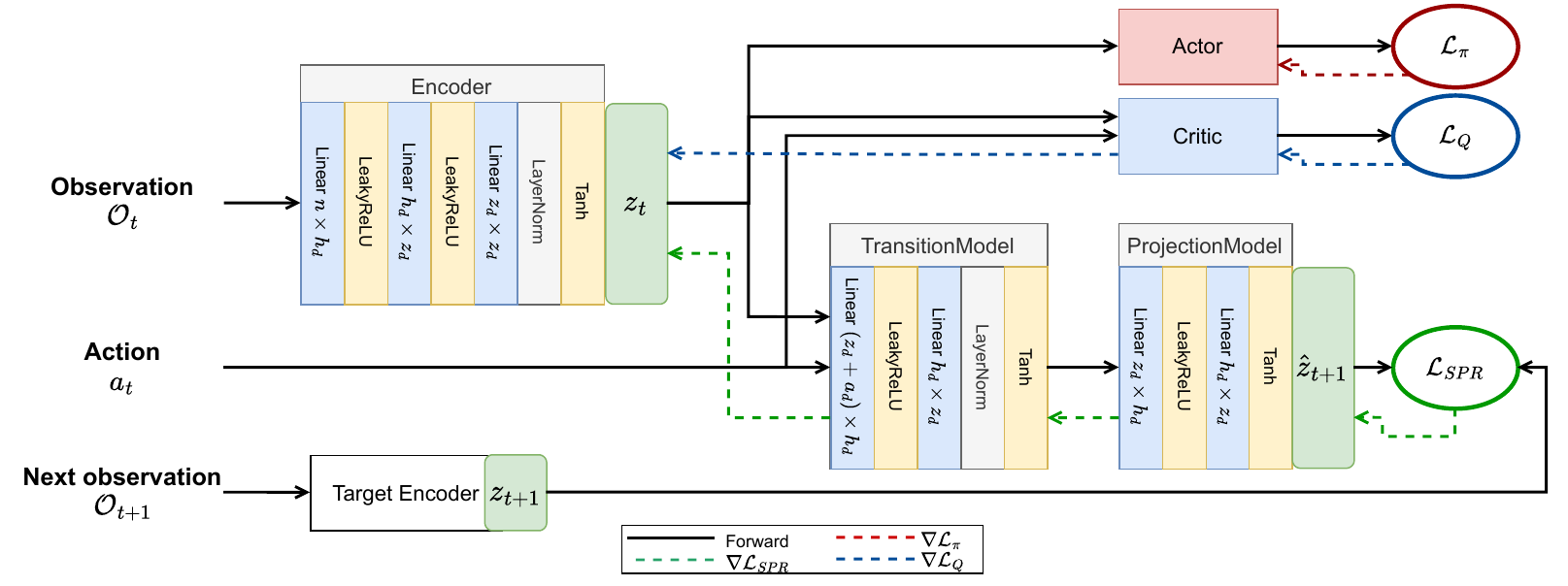}
    \caption[AmelPredDet architecture with a deterministic encoder function.]{AmelPredDet architecture with a deterministic encoder function.}
    \label{fig:SPR_TD3}
\end{figure*}

AmelPredDet is depicted in Figure~\ref{fig:SPR_TD3}, which utilizes an actor-critic algorithm. However, AmelPredDet can also be used with DQN or any other value-based algorithm by replacing the critic function with the $Q$-values estimation function.
AmelPredDet comprises a three-layered MLP as the encoder and a two-layered MLP for the transition and projection model.
For a smooth learning process, the LeakyReLU activation was used between all layers, except for the last one.
In the case of the encoder, a LayerNorm and Tanh process the final layer to produce the latent code $z_t$.
Next, the transition model processes the current code $z_t$ and current action $a_t$ with a LayerNorm and Tanh activation on top.
The projection model processes the output with a Tanh activation to infer the next latent code $\hat{z}_{t+1}$ bounded $\in [-1, 1]$.
Finally, the target encoder model produces the true next code $z_{t+1}$ from the next observation $o_{t+1}$.
As a deterministic encoder function, AmelPredDet optimization used Info Noise Contrastive Estimation (InfoNCE)~\cite{oord2018representation} as a loss function, defined by:
\begin{equation}\label{eq:infonce_loss}
    \mathcal{L}_{SPR} = -\underset{X}{\mathbb{E}}
            \Bigg[\log \frac{f_k(x_{t+k}, ct)}{\sum_{x_j \in X} f_k(x_j, ct)} \Bigg].
\end{equation}

The stochastic version, AmelPredSto, 
which similar to its deterministic version, can be used with actor-critic and value-based RL algorithms.
The main differences between AmelPredSto and AmelPredDet lie in the encoder and projection model, which now uses a probability representation of the latent code.
The encoder model duplicates the number of units in the output to infer the mean $\mu$ and standard deviation $\sigma$ of a multivariate normal distribution.
Next, a Tanh activation bound the latent representation $\in [-1, 1]$.
Then, the projection model, similarly to the encoder, produces a probability distribution without the Tanh activation on top.
Finally, as both $z_{t+1}$ and $\hat{z}_{t+1}$ are probabilities, the Kullback-Liebler divergence (KL)~\cite{csiszar1975divergence} was used for AmelPredSto's optimization, formalized as:
\begin{equation}
     \mathcal{L}_{SPR} = D_{\text{KL}}(P||Q) = \sum_{x \in X}P(x)\log\frac{P(x)}{Q(x)}.
\end{equation}

The encoder function of AmelPredDet and AmelPredSto, is jointly optimized with the $Q$-value estimations.
Hence, in Figure \ref{fig:SPR_TD3} the black dense arrow depicts a forward pass, while the dashed colored lines depict the backward pass.
The $Q$'s backward pass from $\mathcal{L}_Q$, in blue, affects both its model and the encoder model.
The representation backward pass from $\mathcal{L}_{SPR}$ in color green, affects the encoder, transition, and projection models in an end-to-end manner.
The target encoder is a copy of the encoder model updated with Polyak averaging.


\section{Sim2real object-goal scenario}\label{section:environment}
The environment comprises a Crazyflie mini drone\footnote{https://www.bitcraze.io/products/crazyflie-2-1-plus}, with a $2.3 \times 2.3 \times 2.0$ meters delimited flight area, in $ x$, $ y$, and $z$ coordinates.
The UAV's main objective is to reach a goal pose near the target location.
The entire available flight area was divided into a grid with $3 \times 3 \times 3$ cells, using the center of edge cells' location to be a target location.
Hence, the scenario comprises 24 different target locations with eight target quadrants on three different height level $0.5$, $1.025$, and $1.55$ meters of altitude.
The OGN problem is considered solved when the UAV reaches a safe goal distance and looks towards the objective at the same height.

The MDP definition for OGN begins with the action space declaration, which represents the UAV's local velocity.
The UAV's base action space was in a continuous domain as:
\[A = \{\dot{x}, \dot{y}, \dot{z}, \dot{\psi}\},\]
with corresponding translational and rotational velocities as control commands defined by: $\dot{x}, \dot{y}, \dot{z} \in [-0.5, 0.5] \frac{m}{s}$ and $\dot{\psi} \in [-72, 72] \frac{deg}{s}$.
An additional discrete domain was also defined, using the upper and lower limits of each action component plus a no-action case, totaling nine actions.
In this regard, the RL agent can perform a full quadcopter motion in any direction in 3D space.

Next, the observation space used two different settings with varying components regards the target location.
All settings comprise the common UAV control state as base observation, defined by:
\begin{equation}\label{eq:obs_base}
  \mathcal{O}_{base} = \{\theta, \phi, \psi, \dot{\theta}, \dot{\phi}, \dot{\psi}, x, y, z, \dot{x}, \dot{y}, \dot{z}\},
\end{equation}
where $x, y, z$ are the position in the North-East-Up coordinate system, and $\dot{x}, \dot{y}, \dot{z}$ its respective translational velocities.
$\phi, \theta, \psi$ are the roll, pitch, and yaw local Euler angles, and $\dot{\phi}, \dot{\theta}, \dot{\psi}$ their respective angular velocities.
The first observation setting then extends $\mathcal{O}_{base}$ by including the target's raw-coordinates as:
\begin{equation}\label{eq:obs_TC}
  \mathcal{O}_{TC} = \{\mathcal{O}_{base}, \eta, T_{x,y,z}\},
\end{equation}
where $T_{x,y,z}$ are the cartesian values for the target location in a 3D space, and $\eta$ is a north reference towards the target in radians.
In the second observation setting, the target coordinates are replaced by sensor-based information to perceive where the target is, defined by:
\begin{equation}\label{eq:obs_TS}
  \mathcal{O}_{TS} = \{\mathcal{O}_{base}, \text{TS}_{1..6}\},
\end{equation}
where $\text{TS}_{1..6}$ is the information from six virtual sensors, computed using current distance, orientation, and altitude differences with respect to the target's position.
Each virtual sensor represents the activation located on the back, left, and right sides, and three others located in front of the quadcopter.
All the components in the observation space are normalized $\in [-1,1]$.

The reward function encodes the OGN problem to be solved, delivering a value conditioned by the goal distance $D_g$ and the distance threshold $D_{\text{thr}}$.
$D_\text{thr}$ is used with a risk distance value $D_\text{risk}$ as an offset for $D_g$.
The overall reward function comprises distance, orientation, pose, and velocity factors, which are defined below.
\begin{itemize}
  \item The distance component is used to ensure a minimum space between the UAV and the target location, preventing damage to the quadrotor, formalized as:
  \[D_g = D_\text{risk} + \frac{D_\text{thr}}{2},\]
  \begin{equation}\label{eq:r_dist}
      r_{\text{dist}} = - |1 - \frac{DT_t}{D_g}|, \ \ r_{\text{dist}} \in (-\infty, 0],
  \end{equation}
  where $DT_t$ is the Euclidean distance between the UAV and the target at moment $t$.
  
  \item The orientation component used the absolute angle difference between the UAV and target orientation, $\text{UAV}_\mathrm{ori}$ and $\text{T}_\mathrm{ori}$, respectively, to define if it is looking towards the target by:
\begin{equation}
  r_{\text{ori}} = -\Bigg|\frac{\text{UAV}_\text{ori} - \text{T}_\text{ori}|}{\pi}\Bigg|, \ \ r_{\text{ori}} \in[-1, 0].
\end{equation}

  \item An elevation component appears to define the performance with respect to the height difference $\Delta z$ and x-y axes distance $D_{x,y}$, between the UAV and the target as:
  \begin{equation}  
    r_{\text{elev}} = -\Bigg|\frac{\arctan(D_{x,y}, \Delta z)}{\pi / 2}\Bigg|, \ \ r_{\text{elev}} \in[-1, 0].
  \end{equation}

  \item The pose term is updated to consider also the elevation term, defined as:
  \begin{equation}
    r_{\text{pose}} = r_{\text{dist}} + r_{\text{ori}} + r_{\text{elev}}, \ \ r_{\text{pose}} \in (-\infty, 0].
    \end{equation}
  
  \item Finally, the velocity component uses the derivative of the distance $d$ with unbounded values, formalized as:
  \begin{equation}
    r_{\text{vel}} = \frac{d_t - d_{t+1}}{\Delta t},
  \end{equation}
with $\Delta t$ being the actual time difference between moments $t$ and $t+1$.
\end{itemize}

A $r_\text{penalty}$ score discourages the agent with a reward signal $-2$ when it goes out of the allowed area, senses a close object, or surpasses the risk zone.
Or if the agent somehow turns downwards or remains still for five simulated seconds, the run is ended.
When it senses an object but not too close, the penalty value is $-1$.
A success reward value $r_{\text{success}} = 10$ is obtained when the UAV reaches the goal pose or $0$ otherwise.
The success case is when the agent reaches a goal pose defined by:
\begin{equation}
  (1 + r_{\text{dist}}) \times (1 + r_{\text{ori}}) \times (1 + r_{\text{elev}}) > 0.95,
\end{equation}
which indicates if the agent solved the navigation task and must end the run.
Hence, the total step reward is formalized by:
\begin{equation}
  r_{\text{total}} = r_{\text{vel}} + r_{\text{pose}} \times 0.1 + r_{\text{penalty}} + r_{\text{success}}.
\end{equation}

\subsection{Simulated environment}
\begin{figure}
    \centering
    \includegraphics[width=0.95\columnwidth]{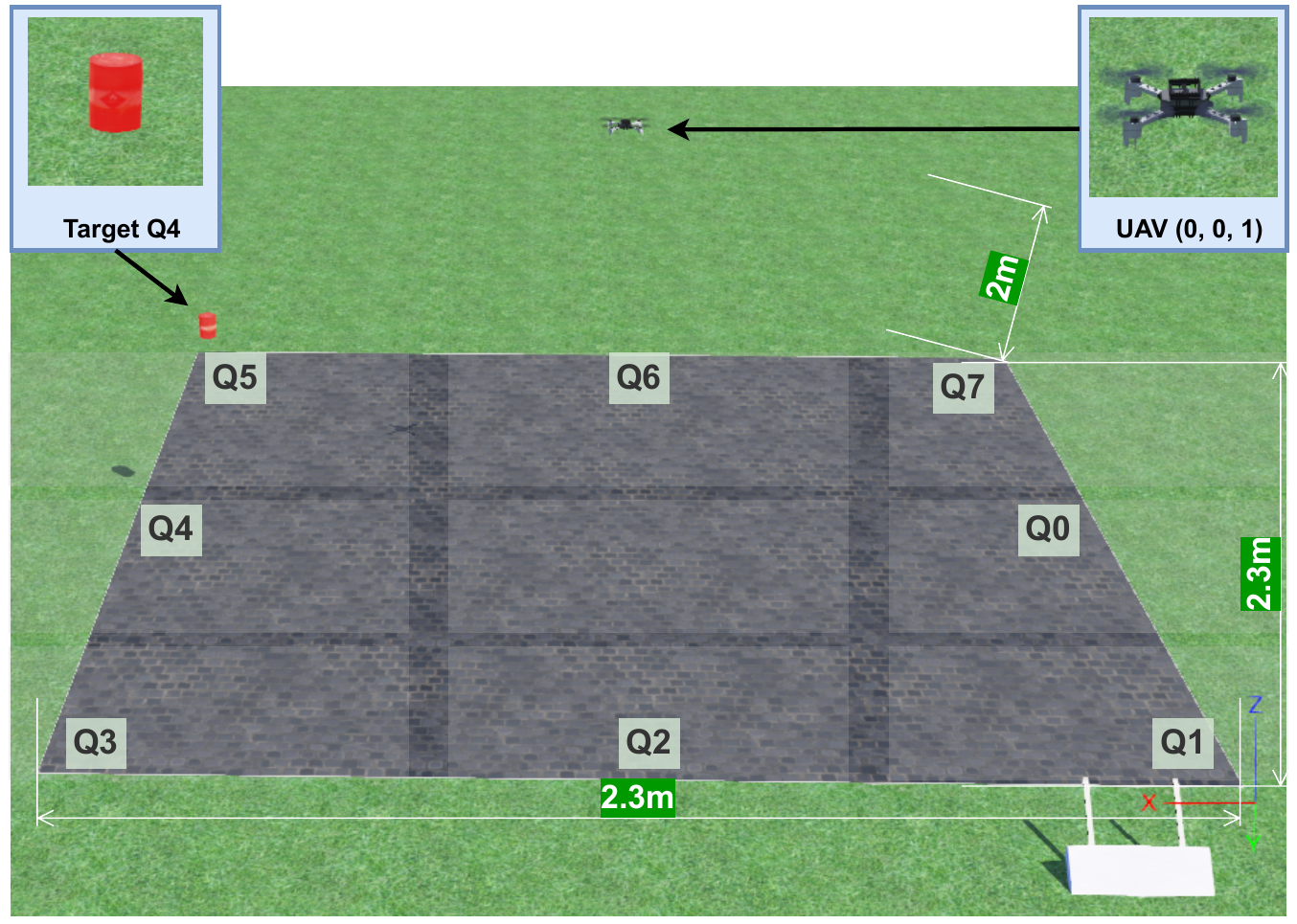}
    \caption[The simulated OGN scene.]{The simulated OGN scene comprises an outdoor landscape with a flight area of $2.3 \times 2.3 \times 2$ meters, a Crazyflie quadcopter, and a red barrel as the target object.
    The target can appear in 24 different locations on three levels, each level comprises 8 grids at 0.5m height, as depicted above.
    The second and third levels are located at $1.025$ and $1.55$ meters height, respectively.
    At each episode, the agent starts from the same initial position at $(x=0, y=0, z=1)$.}
    \label{fig:environment_cf}
\end{figure}

The simulated UAV scene was projected in an outdoor landscape, using the quadcopter, and the target object.
The scene only includes a red barrel as a target object for visual cues during evaluation.
At the beginning of each episode, the target location can randomly appear in certain quadrants.
Hence, the expected trajectory is a straight line from the initial position to the goal pose near the target location.
Additionally, while the UAV is in transit, it should adjust its orientation and altitude accordingly to reach the goal pose near the target.
A delimited flight area is also used to prevent the UAV from flying out of range, which is considered the remote control signal distance.
The original Crazyflie PID's firmware was used as a low-level controller, which simplifies further real-world transition.
Figure~\ref{fig:environment_cf} depicts eight target quadrants on the first level at $0.5$ meters height.
A public GitHub repository hosts the simulated Webots environments\footnote{https://github.com/angel-ayala/gym-webots-drone}.

\subsection{Real-world settings}
\begin{figure}
    \centering
    \includegraphics[width=0.95\linewidth]{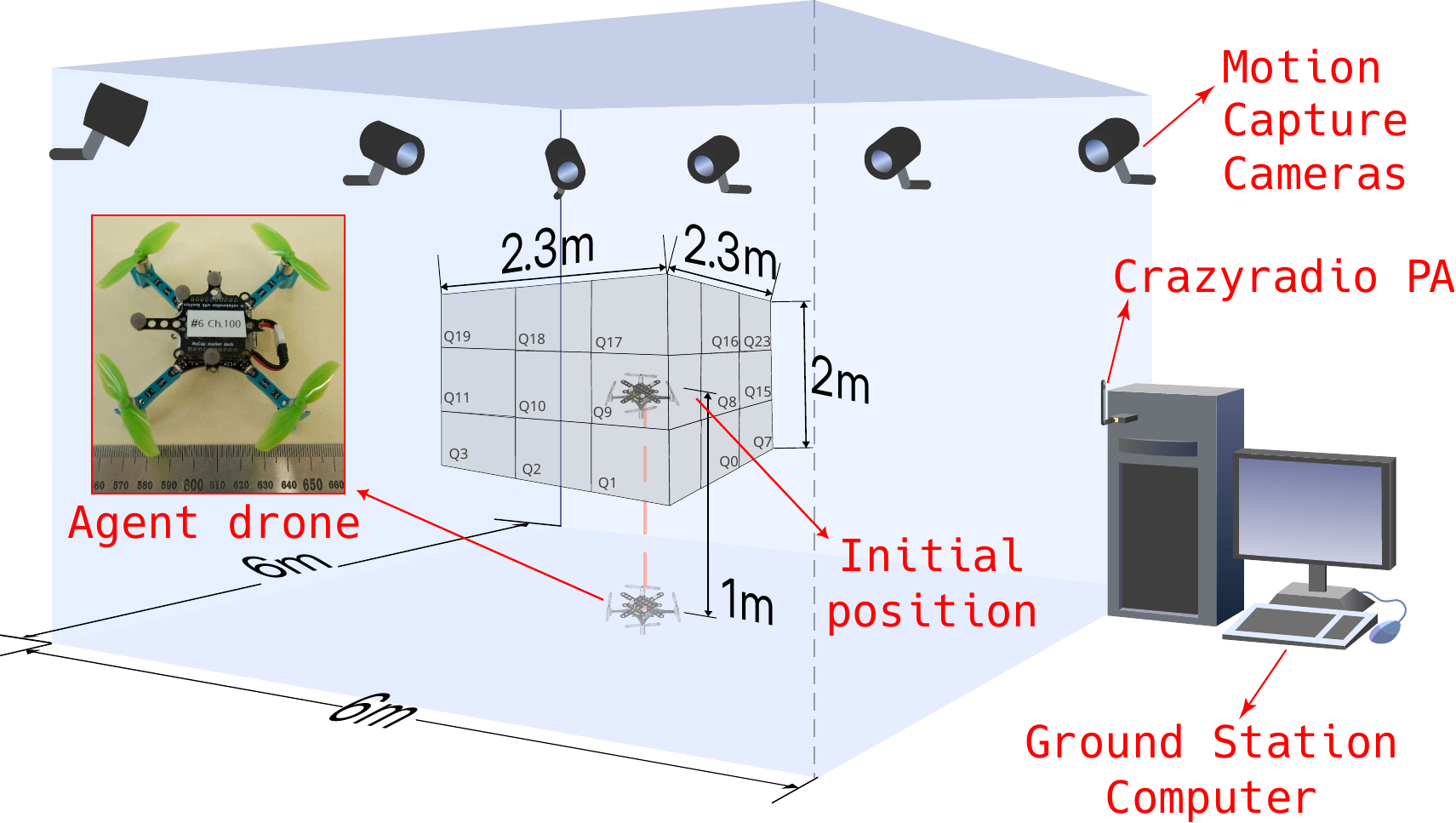}
    \caption[The real-world OGN setup.]{Real-world flying arena for autonomous quadcopter setup.
    A $6 \times 6$ meters room comprises a serie of OptiTrack motion capture cameras for reading the position and rotational information through the reflection of infrared marker on a Crazyflie drone.}\label{fig:real-world_setup}
\end{figure}

The real-world counterpart comprises a flying arena for autonomous aircraft.
The arena comprises a ground station connected to a series of OptiTrack motion capture cameras in a $6 \times 6$ meters room.
Nevertheless, due to the camera's field of view on the reflection of infrared marker, a reduced space of $2 \times 2$ is available for safe flights.
Additionally, vertical limits are also needed, from a minimum altitude of 0.5 until 2 meters.
A representation of the real-world setup is presented in Figure~\ref{fig:real-world_setup}.
Real Crazyflie velocities constraint was also considered to prevent control signals from differing too much in magnitude.
In this regards, sensor readings are delivered asynchronously by different components on the flying arena.
For example, the position and rotational information were obtained from the Motive software of the motion capture system, with respective velocities computed from their derivatives.
Nevertheless, roll, pitch, and yaw angles and battery level values were obtained from the Crazyflie platform.
In this regard, the twelve variables for $\mathcal{O}_{base}$ are available to define the quadcopter state.

\section{Experimental Navigation Results}\label{section:results}

The experimental settings addressed the comparison of the proposed SRL perception module named Amelkantun Prediction (AmelPred) with different observation space and RL algorithms.
Regarding the observation, the raw target coordinates $\mathcal{O}_{TC}$ and a sensor-based target observation $\mathcal{O}_{TS}$ were used.
We selected DQN, TD3, and SAC as baseline RL algorithms and evaluated the efficiency improvements achieved with the proposed AmelPred SRL method.
The development of the AmelPred architecture was built upon the Stable-Baselines3 library.

Five different seed values were used for each method to learn a suitable navigation policy.
As off-policy RL algorithms, the optimization process used a random experience replay buffer with $2^{16}$ ($\sim 65K$) transitions.
The first model parameter optimization began after $2^{11}$ ($\sim 2K$) steps for memory initialization.
The training process consisted of $450K$ steps, chunk into episodes of $9K$ steps each.
Default SB3 learning hyperparameters were used, such as Adam learning rate of $1 \times 10^{-4}$ for DQN, $3 \times 10^{-3}$ for SAC, and $1 \times 10^{-3}$ for TD3.
The target network parameters update was achieved through Polyak averaging using $\tau = 5 \times 10^{-3}$, as compared to its original counterpart.
Only DQN algorithm used $\tau = 1$, updated every $10K$ steps.
All agents received an observation as input, with a random number of steps $\in [5, 7]$ between consecutive observations.

All variants of the AmelPred representation model were optimized using Adam with a learning rate of $1 \times 10^{-3}$.
Both loss values from the $Q$-value estimations and the representation models were used for joint optimization of parameters.
The target encoder network was also updated through Polyak averaging with $\tau = 0.999$.
In the case of TD3-AmelPred, the critic and actor models match the SAC architecture, featuring two layers of MLP with 256 units each.
The TD3 architecture reduction was the only change made; all the other aspects remain untouched.
The training parameters were used for both main AmelPredDet and AmelPredSto variations.

The evaluation phase was executed for each trained agent checkpoint at episodes 5, 10, 20, 35, and 50, with 240 iterations each.
Additionally, experimental results included two other SRL methods from the literature.
The first method was a modified version of SPR~\cite{schwarzer2020data}, which follows a similar architecture proposed by the authors.
The original implementation of SPR was used in conjunction with Rainbow for pixel-based observations, whereas this study presents a modified version of SPR for handling vector-based observations.
The second method was an unmodified baseline for SRL presented by \cite{ni2024bridging}, which comprises the TD3 algorithm and three different techniques for target model updates and loss computation.
Among all the technique combinations presented, the L2 loss with the online target model was the only one that showed convergence, named here as TD3-Ni.
All RL performance-related metrics were obtained using the RLiable Python library with normalized scores using the TD3 maximum reward value.

\subsection{Simulated experimental performance}

\begin{figure}
    \centering
    \includegraphics[width=0.9\linewidth]{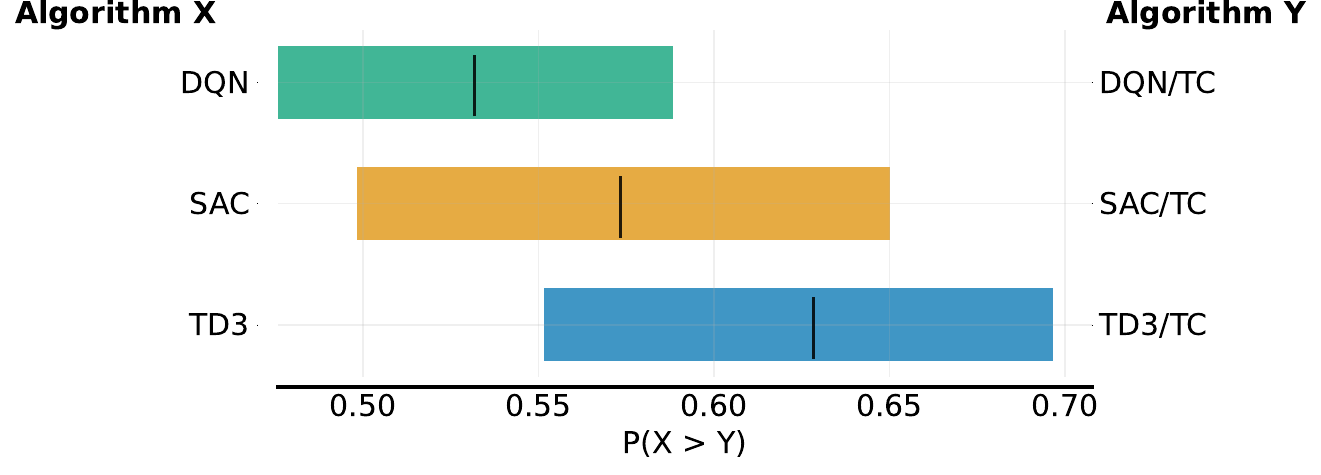}
    \caption[Probability of improvement between vanilla algorithms using target sensing and target coordinates in the agent's observation.]{Probability of improvement between vanilla algorithms using target sensing and target coordinates in the agent's observation, with five different seed values.
    The last ones are denoted with /TC.}\label{fig:TS_TC_probability}
\end{figure}

The first experimental results aimed to compare the performance enhancement from replacing the target location in each agent's observation with a sensor data-related approach.
The algorithms labeled as DQN, SAC, and TD3 use the target sensing approach.
In contrast, the DQN/TC, SAC/TC, and TD3/TC labeled algorithms use raw target coordinates.
The comparison in terms of probability of improvements between each vanilla algorithm is presented in Figure~\ref{fig:TS_TC_probability}.
The probability of improvement indicates the likelihood of achieving a higher reward value for algorithm X compared to algorithm Y.
In the case of TD3, it is the algorithm that performs better than TD3/TC with a 64\% probability.
Then, the SAC algorithm has only a 56\% probability of achieving a better reward than SAC/TC.
Finally, in the case of DQN, it is the algorithm that slightly improves DQN/TC with a 54\% probability.
The obtained results indicate that using a target sensing approach is more efficient than using target coordinates on all vanilla algorithms.

Hence, next experimental results were focused on comparing the RL vanilla algorithms against the RL algorithm with AmelPred SRL method.
Additionally, for comparison against the state-of-the-art, the unmodified TD3-Ni and the adapted SPR methods were included in the presented results.
The first method, named TD3-Ni, utilizes the TD3 algorithm with an L2 distance from the latent space and an online target model for representation loss~\cite{ni2024bridging}.
The second adapted SPR method is a modified approach designed to be compatible with vector-based observations~\cite{schwarzer2020data}.

\begin{figure}
    \centering
    \includegraphics[width=0.95\columnwidth]{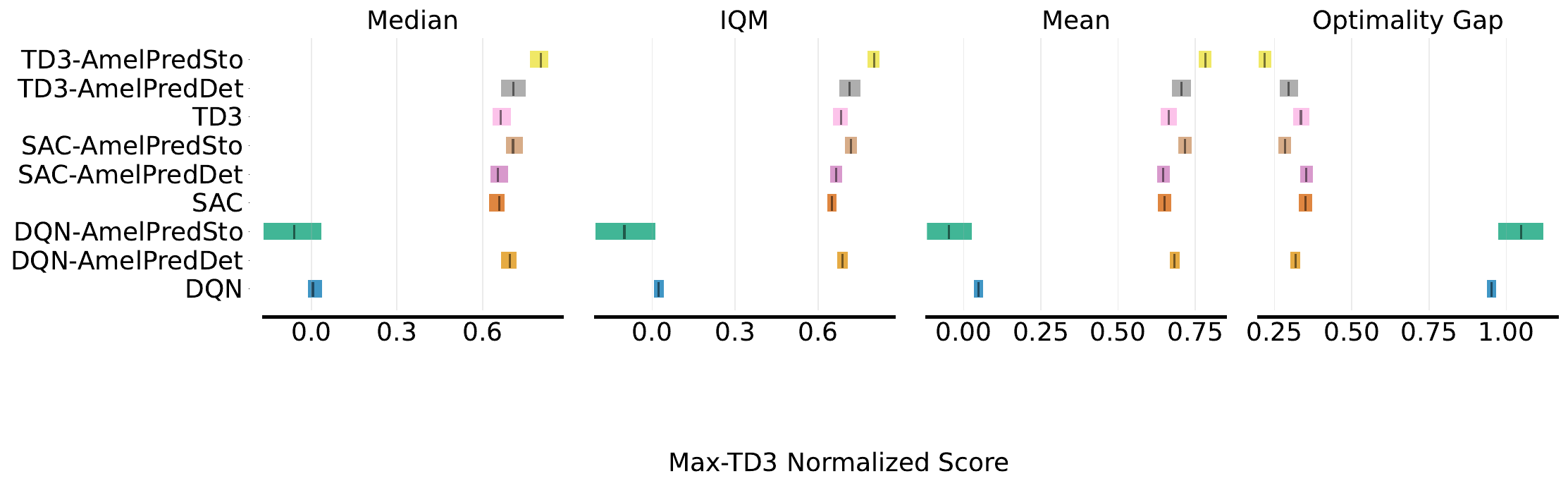}
    \caption{Aggregated reward values for comparing vanilla algorithm and its AmelPred version, with five different seeds value.}
    \label{fig:TS_reward_agg}
\end{figure}

The first analysis compares the obtained reward scores of each vanilla algorithm against the AmelPredDet versions, depicted in Figure~\ref{fig:TS_reward_agg}.
A remarkable outcome was achieved by DQN-AmelPredDet, which increases the reward score by 6.5 times compared to DQN, surpassing SAC and achieving similar rewards than TD3.
In the case of actor-critic algorithms, only TD3-AmelPredDet outperforms TD3 in all presented reward values.
In contrast, SAC-AmelPredDet achieves nearly the same reward as SAC, with no notable difference.
Therefore, the presented AmelPredDet proposal works fine with deterministic algorithms such as TD3 and DQN.

A second analysis compares the AmelPredSto state representation approach against the AmelPredDet approach.
As initially thought, it was expected for AmelPredSto be compatible with SAC algorithm given the same stochastic nature.
However, TD3-AmelPredSto achieved the best reward value in comparison to all other methods, obtaining a 0.8 normalized IQM reward value.
In second place was the SAC-AmelPredSto method, which achieved better rewards than SAC-AmelPredDet, demonstrating the compatibility of the stochastic processes between AmelPred and the SAC algorithm.
Unlike TD3, the DQN-AmelPredSto method underperforms DQN, whereas the DQN-AmelPredDet methods outperform it, possibly due to the bootstrapped way in computing the targets for $Q$-values of DQN is more stable with deterministic values.

\begin{figure}
    \centering
    \includegraphics[width=0.95\columnwidth]{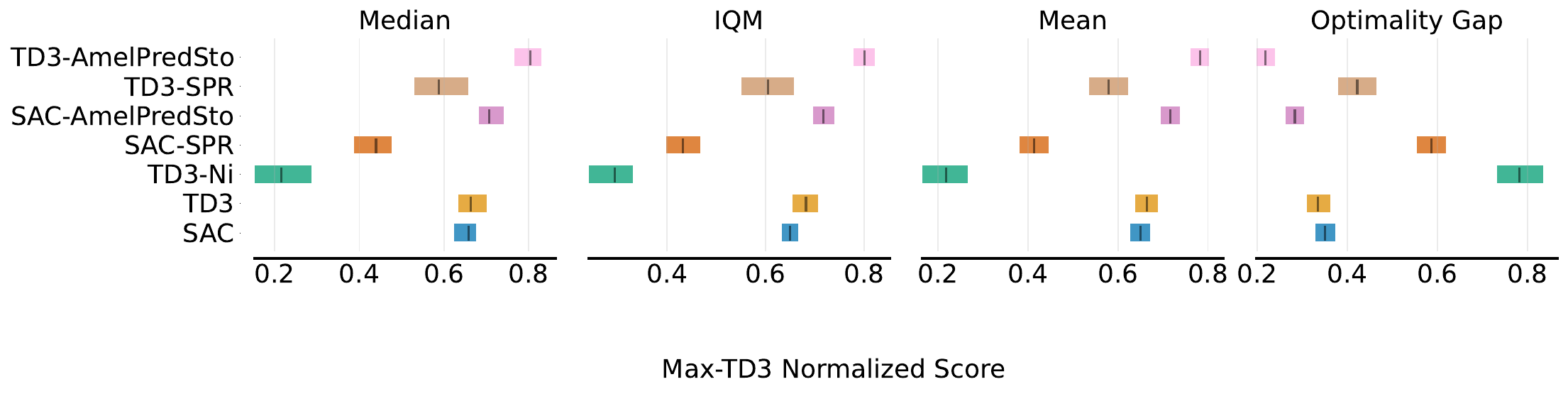}
    \caption{Aggregated reward values for comparing the AmelPredSto approach against literature methods, with five different seeds value.}
    \label{fig:TS_sota_rewards_agg}
\end{figure}

The third analysis from Figure~\ref{fig:TS_sota_rewards_agg} was to compare the performance against state-of-the-art (SOTA) methods.
In this regard, only SAC and TD3 algorithms were considered in an attempt at fair comparison against the TD3-Ni method.
From the observed results, TD3-AmelPredSto again surpasses all presented methods, with the 0.8 normalized IQM reward value.
In second place was the SAC-AmelPredSto method with a normalized IQM reward value of 0.7.
The vanilla TD3 and SAC algorithms achieved third and fourth place, respectively.
The included SOTA methods were not able to solve the object-navigation problem, achieving lower reward values than SAC.
The fifth and sixth places were given to TD3-SPR and SAC-SPR, respectively, achieving IQM reward values of 0.45 and 0.6, respectively.
The final place was for the TD3-Ni method, which achieved a 0.2 IQM reward value.

In terms of computational cost, as presented in Table~\ref{tbl:TS_complexity}, each method is compared based on the number of floating-point operations (Flops) and the number of parameters.
An overall comparison between the algorithm types shows that the DQN-based methods are the most lightweight during both the learning and evaluation stages, followed by the SAC and TD3 algorithms.
Additionally, during evaluation, all algorithms comprise fewer floating-point operations and parameters, since the target models are not considered.
As state representation learning approaches comprise additional models, it is expected that those methods will increase the model's complexity.
For example, when evaluating the best models, it can be observed that DQN-AmelPredDet performs almost 6 times more Flops than DQN.
Between SAC-AmelPredSto and SAC, the difference is minor, with the first one comprising 1.2x more Flops than the last.
In the case of TD3-AmelPredSto, a complexity reduction of 1.5 times that of TD3 was achieved, primarily due to the reduction in the number of parameters of the policy model.

\begin{table}
    \centering
    \caption{Computational cost comparison using sensor-based target-perception in observation.}
    \begin{tabular}{l|cc|cc}
        \toprule
        & \multicolumn{2}{c|}{\textbf{Learning}}
        & \multicolumn{2}{c}{\textbf{Evaluation}} \\
        \textbf{Method}
        & Flops & \# Params
        & Flops & \# Params \\
        \midrule
        \textbf{DQN}             &  11.648K &  11.922K &   5.824K &   5.961K \\
        \textbf{DQN-AmelPredDet} & 137.216K & 139.538K &  33.472K &  34.121K \\
        \textbf{DQN-AmelPredSto} & 109.056K & 110.706K &  34.624K &  35.241K \\
        \midrule
        \textbf{SAC}             & 357.888K & 360.460K & 143.616K & 144.649K \\
        \textbf{SAC-SPR}         & 469.248K & 473.420K & 179.584K & 181.129K \\
        \textbf{SAC-AmelPredDet} & 497.024K & 501.644K & 177.536K & 179.081K \\
        \textbf{SAC-AmelPredSto} & 468.864K & 472.812K & 178.688K & 180.201K \\
        \midrule
        \textbf{TD3}             & 773.200K & 777.412K & 257.500K & 258.905K \\
        \textbf{TD3-SPR}         & 542.976K & 547.660K & 178.560K & 180.101K \\
        \textbf{TD3-AmelPredDet} & 570.752K & 575.884K & 176.512K & 178.053K \\
        \textbf{TD3-AmelPredSto} & 542.592K & 547.052K & 177.664K & 179.173K \\
        \bottomrule
    \end{tabular}
    \label{tbl:TS_complexity}
\end{table}

It is expected that SRL methods will improve sample efficiency in RL algorithms, as measured by the reward curve and the performance profile, to compare each presented method.
The first metric proposes assessing the evolution using the normalized interquartile mean (IQM).
Next, the reported performance profile shows the empirical tail distribution function of the reward value.
Both metrics include a shaded area that represents a 95\% pointwise confidence interval based on a stratified bootstrap.
The first analysis compared each algorithm group, followed by a comparison of each presented method, described below.

\begin{figure}
    \centering
    \subcaptionbox{Vanilla comparison.\label{fig:TS_sample_vanilla}}{\includegraphics[width=0.45\columnwidth]{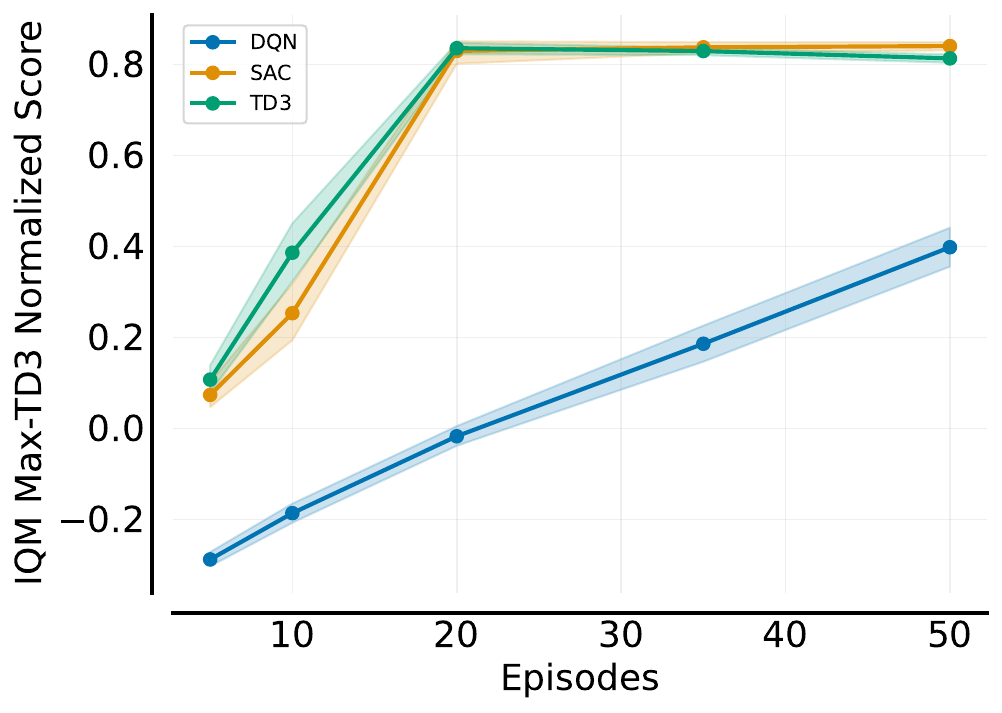}}
    \quad
    \subcaptionbox{AmelPredDet comparison.\label{fig:TS_sample_ispr}}{\includegraphics[width=0.45\columnwidth]{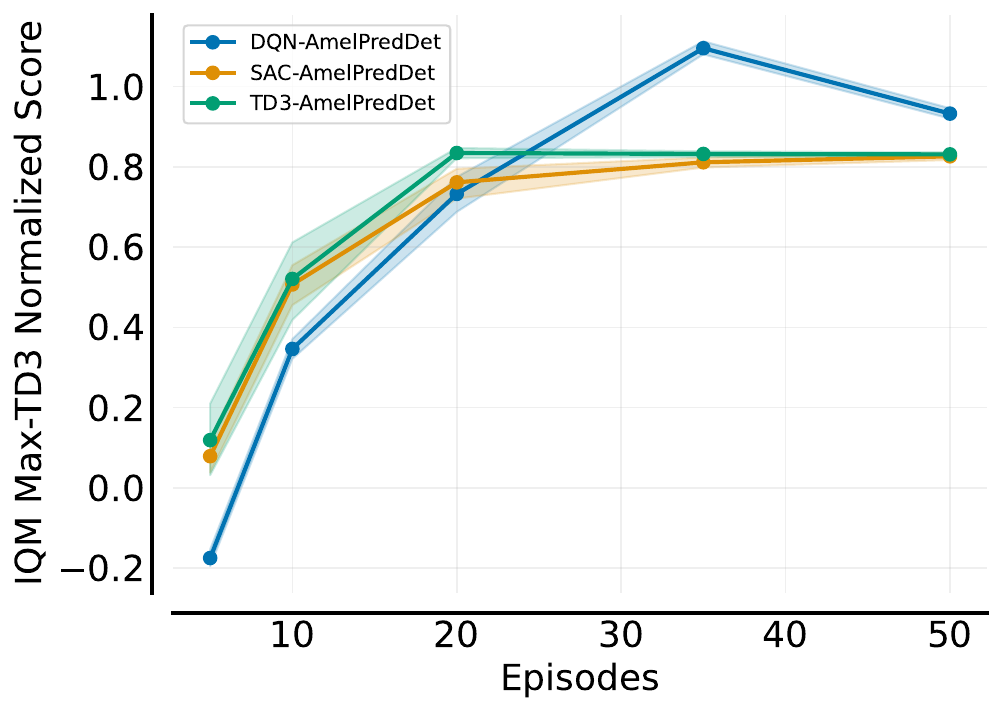}}
    \quad
    \subcaptionbox{AmelPredSto comparison.\label{fig:TS_sample_stch}}{\includegraphics[width=0.45\columnwidth]{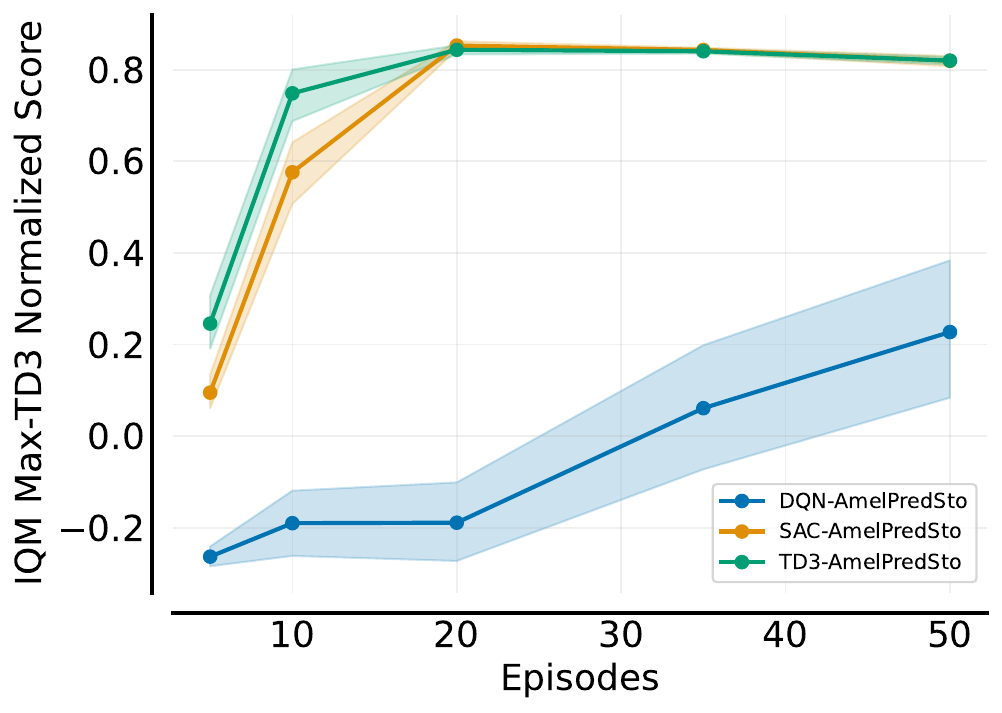}}
    \caption{Interquartile mean (IQM) reward curve comparison of vanilla algorithms and prediction-based methods, with five different seeds value.}\label{fig:TS_sample2}

\end{figure}

Figure~\ref{fig:TS_sample2} presents a comparison between each method type, allowing us to assess the efficiency of each algorithm type.
In the case of vanilla methods in Figure~\ref{fig:TS_sample_vanilla}, TD3 achieved the highest IQM reward values from the earliest episodes, with a decreasing score from episode 20 compared to SAC.
As expected, DQN obtained the lowest IQM scores, requiring more samples to achieve a better outcome, with a constant improvement.
Similarly, the AmelPredDet methods presented in Figure~\ref{fig:TS_sample_ispr}, showed that TD3-AmelPredDet got the highest IQM rewards until episode 20, with SAC-AmelPredDet reaching the same score at the end.
Surprisingly, DQN-AmelPredDet achieved the highest score after episode 20, reaching a 1.2 of IQM in episode 35 and a notable reduction to 0.9 in episode 50.
For the AmelPredSto methods in Figure~\ref{fig:TS_sample_stch}, TD3-AmelPredSto reached almost a 0.8 score in episode 10, followed by SAC-AmelPredSto.
From episode 20 onwards, both SAC-AmelPredSto and TD3-AmelPredSto perform the same.
The stochastic version of AmelPredDet was not compatible with DQN, as DQN-AmelPredSto performed the worst in terms of IQM rewards.

\begin{figure}
    \centering
    \subcaptionbox{Vanilla comparison.\label{fig:TS_performance_vanilla}}{\includegraphics[width=0.45\columnwidth]{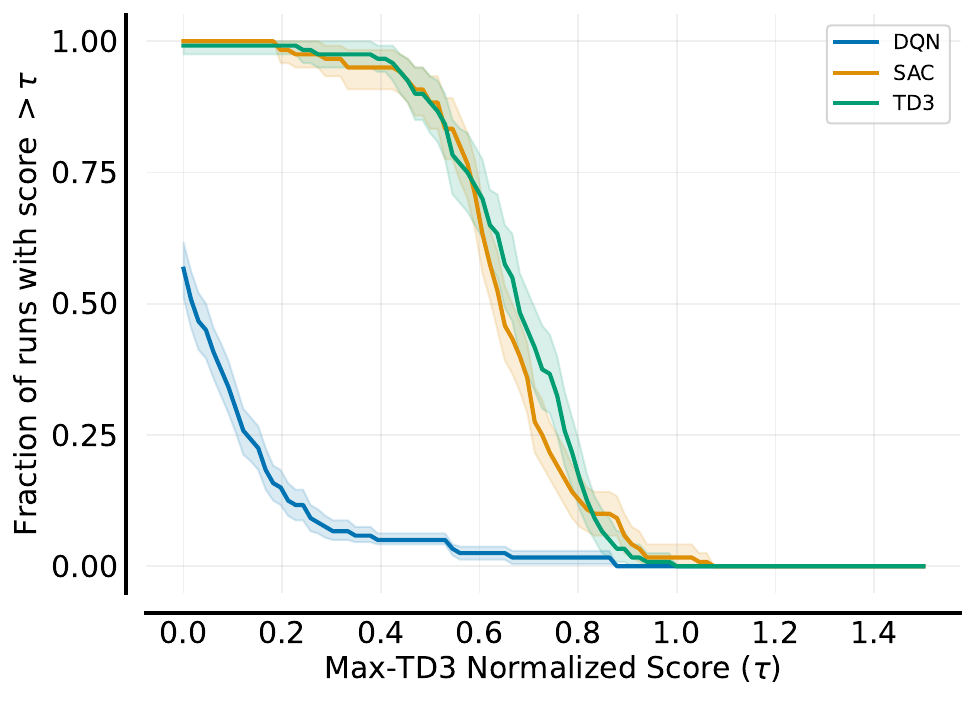}}
    \quad
    \subcaptionbox{AmelPredDet comparison.\label{fig:TS_performance_ispr}}{\includegraphics[width=0.45\columnwidth]{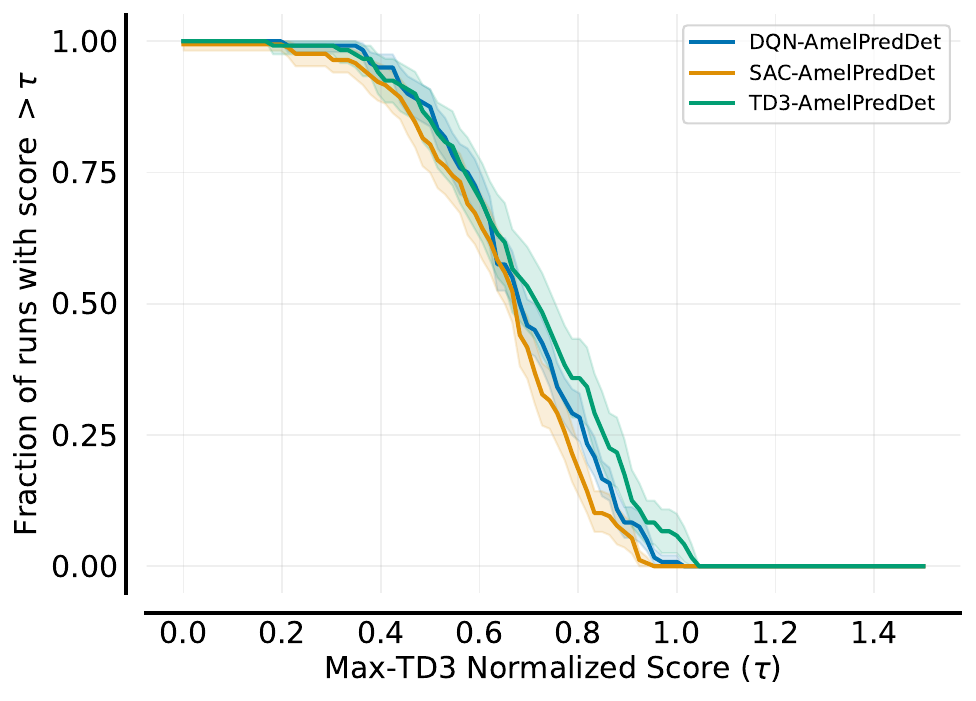}}
    \quad
    \subcaptionbox{AmelPredSto comparison.\label{fig:TS_performance_stch}}{\includegraphics[width=0.45\columnwidth]{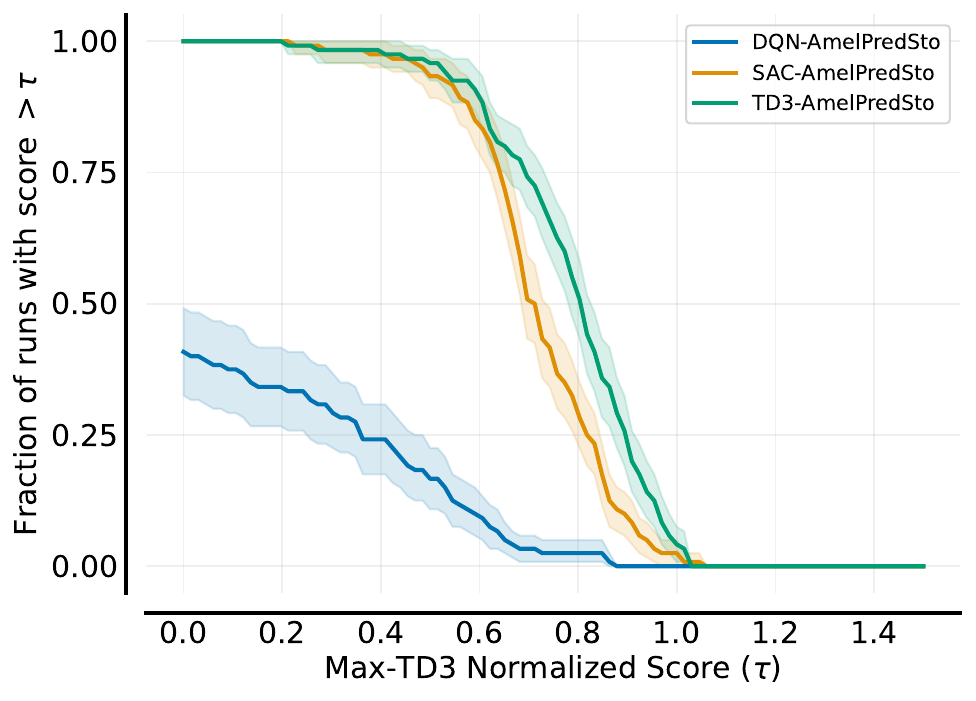}}
    \caption{Performance profile comparison of vanilla algorithms and prediction-based methods.}\label{fig:TS_performance2}

\end{figure}

The performance profiles of the proposed method types are depicted in Figure~\ref{fig:TS_performance2}.
For the case of vanilla methods in Figure~\ref{fig:TS_performance_vanilla}, TD3 and SAC outperform DQN, with TD3 achieving more runs with higher reward values.
In contrast, SAC achieves slightly more runs with scores above 0.8.
Regarding AmelPredDet methods in Figure~\ref{fig:TS_performance_ispr}, all three methods perform similarly, with notable differences in higher reward values.
For example, TD3-AmelPredDet achieved a larger number of runs with rewards above 0.65, followed by DQN-AmelPredDet and SAC-AmelPredDet.
In terms of stochastic AmelPred methods in Figure~\ref{fig:TS_performance_stch}, the results obtained are notably different.
TD3-AmelPredSto performed the best curve, achieving more runs with scores over 0.6, followed by SAC-AmelPredSto.
For DQN-AmelPredSto, it was the worst-performing method, with half of the runs achieving scores below 0.5.

\subsubsection{Sensor-based navigation performance}
The OGN performance assessment was done using the distance to success (DTS), and success path length (SPL) metrics.
DTS, and SPL allowed the assessment of how far the agent ends from the target, and how efficient the trajectory is.
All metrics were obtained by considering a total of 10 trials for each of the 24 target locations, resulting in a total of 240 trials.
The figures presented in this section depict the mean value of five different trained agents, represented by a solid colored line, and their corresponding standard deviation, represented by a shaded colored area.

\begin{figure}
    \centering
    \subcaptionbox{Vanilla comparison.\label{fig:TS_DTS_vanilla}}{\includegraphics[width=0.45\columnwidth]{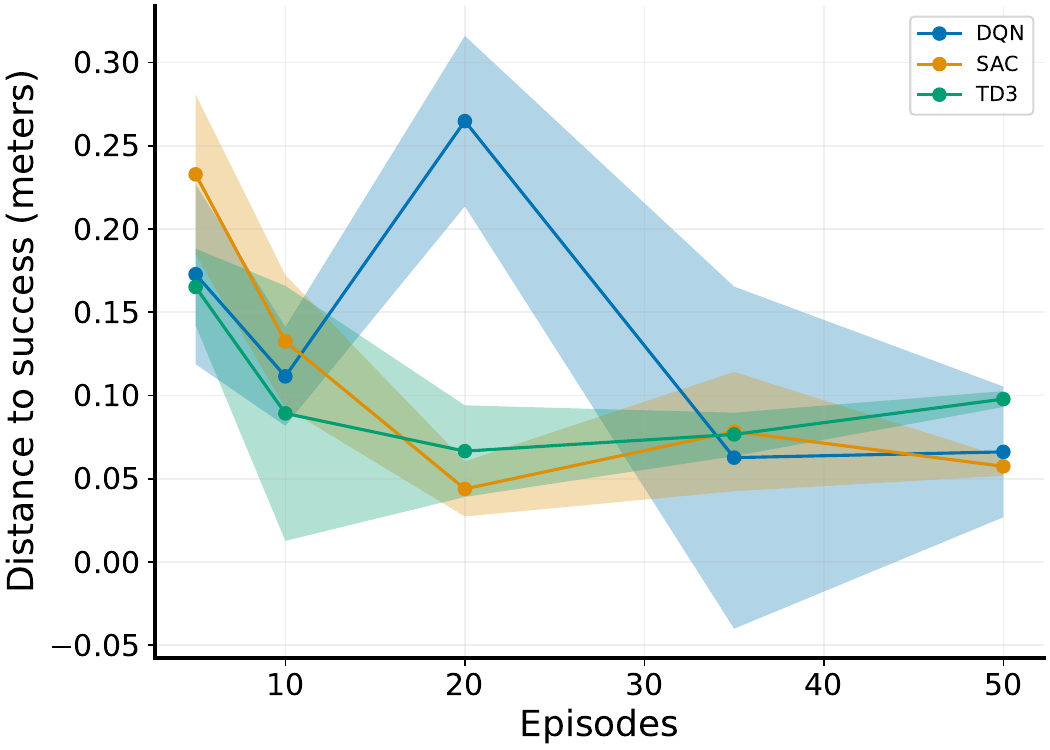}}
    \quad
    \subcaptionbox{AmelPredDet comparison.\label{fig:TS_DTS_ispr}}{\includegraphics[width=0.45\columnwidth]{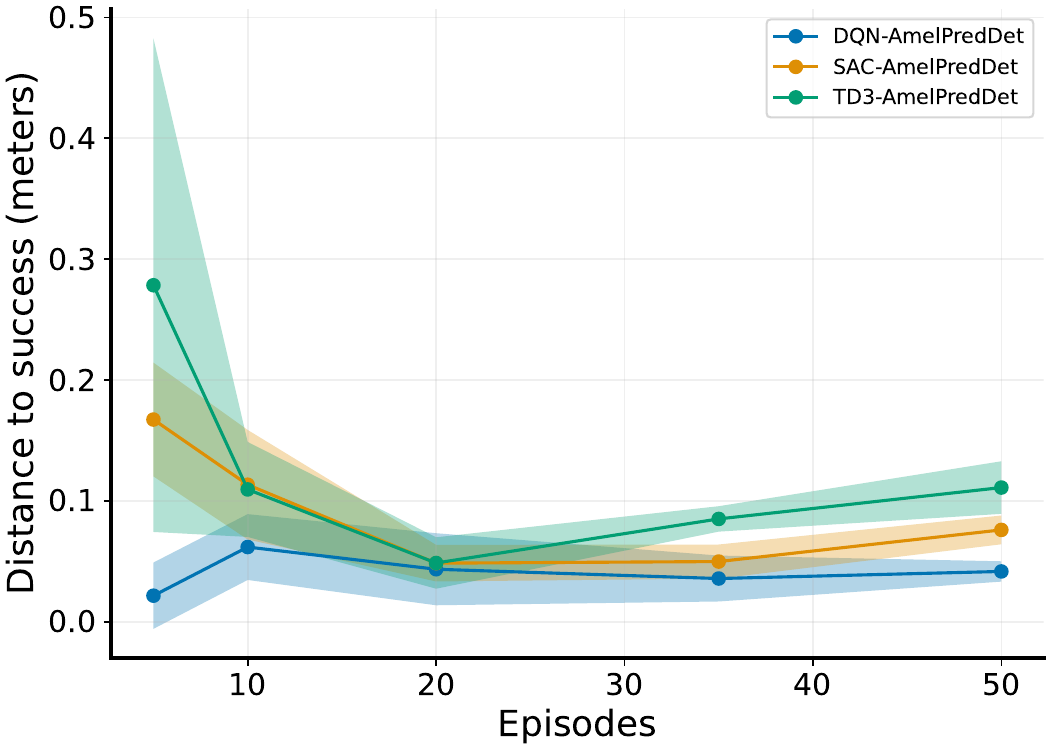}}
    \quad
    \subcaptionbox{AmelPredSto comparison.\label{fig:TS_DTS_stch}}{\includegraphics[width=0.45\columnwidth]{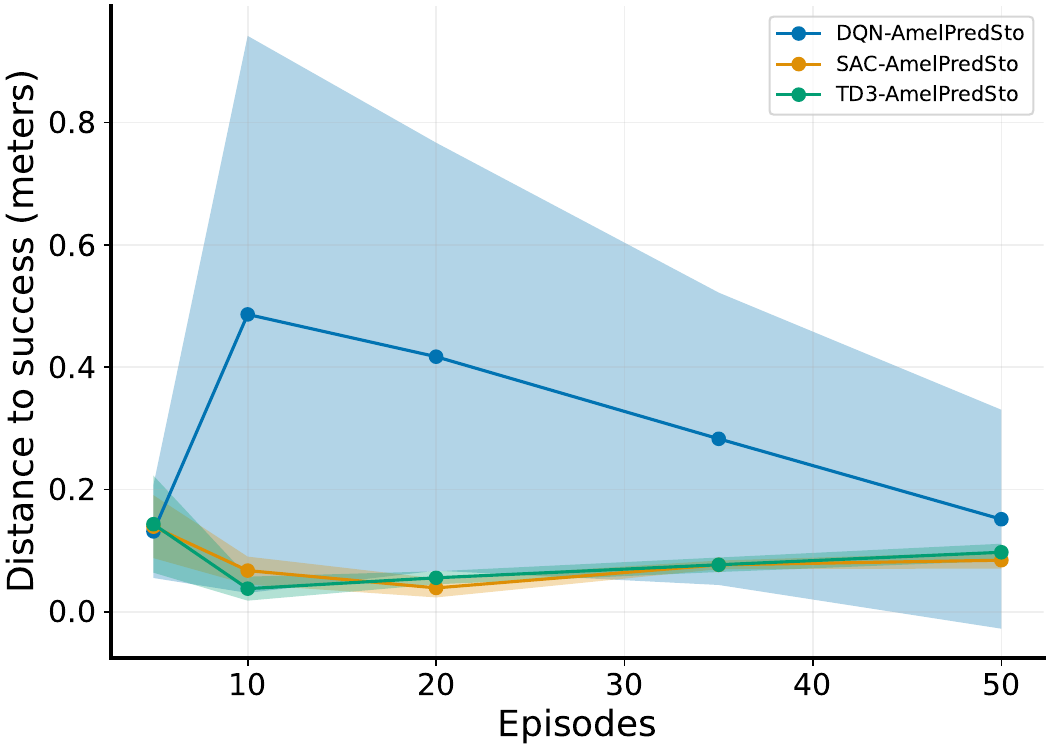}}
    \caption[Distance to success (DTS) comparison of vanilla algorithms and prediction-based methods.]{Distance to success (DTS) metric comparison of vanilla algorithms and prediction-based methods.
    The colored lines depicts the mean value of five agents during evaluation, with the shaded area as the standard deviation.}\label{fig:TS_DTS2}

\end{figure}

Figure~\ref{fig:TS_DTS2} reports the DTS-based comparison across methods.
DQN and TD3 started from a similar distance around 0.17m with a divergent evolution, as shown in Figure~\ref{fig:TS_DTS_vanilla}.
The first one reached a peak of 0.23m and then stabilized around 0.07m from episode 25 onwards.
The last one, which moved to 0.08m at episode 20, ended with a DTS of 0.10m at episode 50.
SAC had the highest initial DTS and the lowest final DTS, with values of 0.23 and 0.06 meters, respectively.
The AmelPredDet methods presented in Figure~\ref{fig:TS_DTS_ispr} showed different DTS curves according to each algorithm.
However, a DTS value of around 0.05m was obtained in episode 20 by all algorithms, diverging in the evolution of their values.
On the one hand, DQN-AmelPredDet had the lowest final DTS, followed by SAC-AmelPredDet and TD3-AmelPredDet.
On the other hand, a remarkable evolution with the proposed AmelPredSto methods can be observed in Figure~\ref{fig:TS_DTS_stch}.
The entire evolution of DTS for the stochastic versions was practically the same, starting and ending with a difference not greater than 0.01m.
A notable difference was observed for DQN-AmelPredSto, which initially increases the distance before reducing it to values around 0.17m.

\begin{figure}
    \centering
    \subcaptionbox{Vanilla comparison.\label{fig:TS_SPL_vanilla}}{\includegraphics[width=0.45\columnwidth]{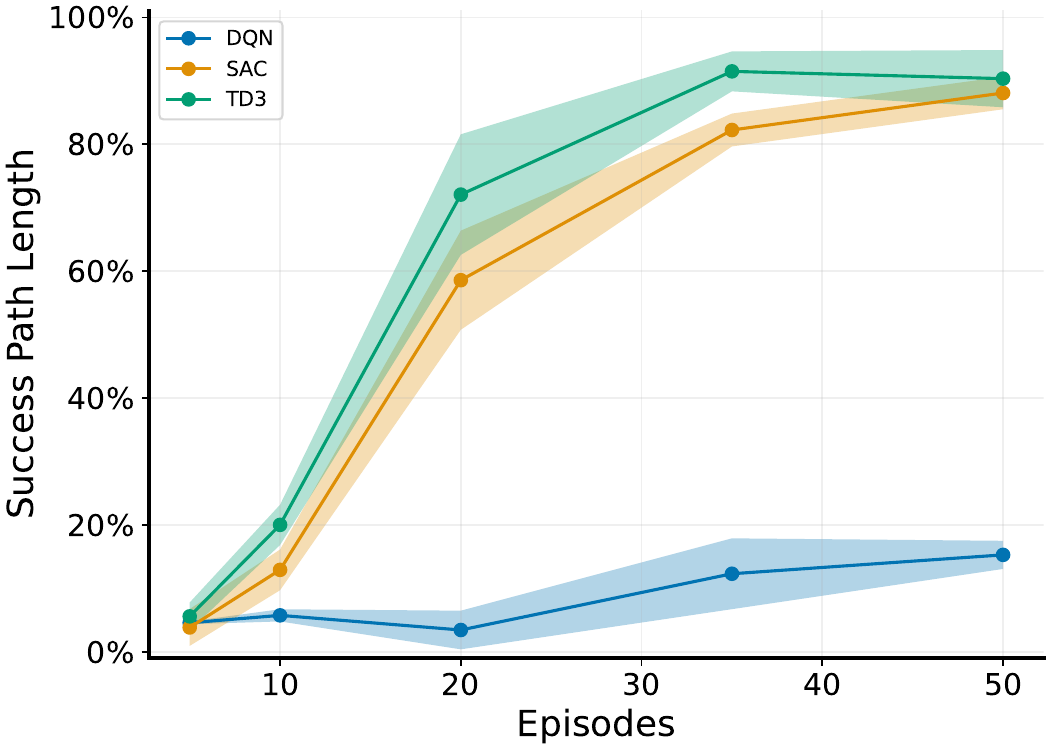}}
    \quad
    \subcaptionbox{AmelPredDet comparison.\label{fig:TS_SPL_ispr}}{\includegraphics[width=0.45\columnwidth]{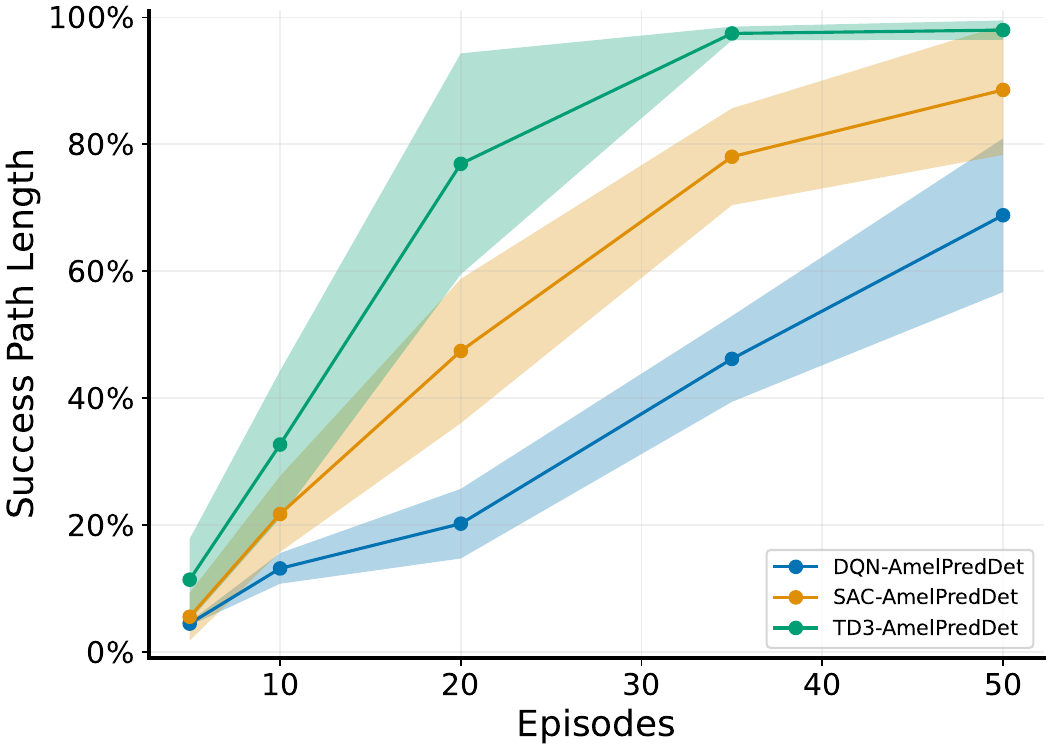}}
    \quad
    \subcaptionbox{AmelPredSto comparison.\label{fig:TS_SPL_stch}}{\includegraphics[width=0.45\columnwidth]{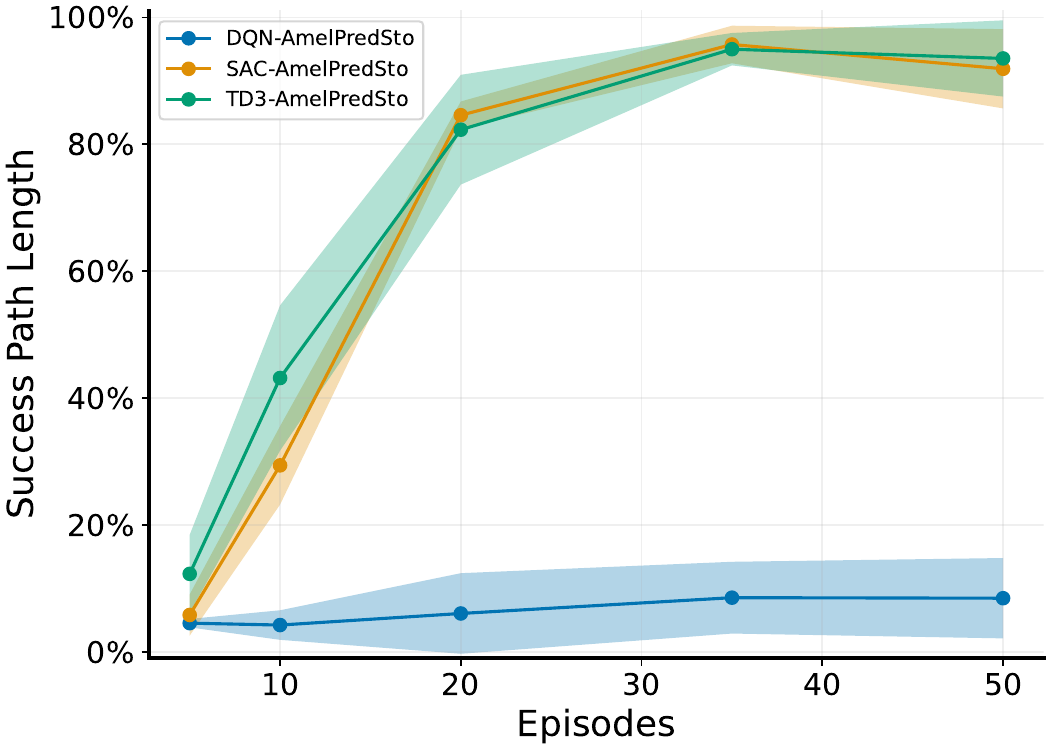}}
    \caption[Success path length (SPL) comparison of vanilla algorithms and prediction-based methods.]{Success path length (SPL) metric comparison of vanilla algorithms and prediction-based methods.
    The colored lines depicts the mean value of five agents during evaluation, with the shaded area as the standard deviation.}\label{fig:TS_SPL2}

\end{figure}

The SPL metric focuses on comparing the achieved trajectory for each algorithm, presented in Figure~\ref{fig:TS_SPL2}.
Vanilla methods, such as TD3, achieved the best curve, followed by SAC, with both reaching a similar SPL of around 90\%, as depicted in Figure~\ref{fig:TS_SPL_vanilla}.
On the other hand, DQN was unable to surpass 20\% of SPL.
A notable gap between AmelPredDet methods can be observed in Figure~\ref{fig:TS_SPL_ispr}, with TD3-AmelPredDet performing the best, achieving 99\% of SPL.
SAC-AmelPredDet and DQN-AmelPredDet demonstrated a consistent improvement over time, reaching final SPLs of 70\% and 90\%, respectively.
A very similar outcome between SAC-AmelPredSto and TD3-AmelPredSto can be observed in Figure~\ref{fig:TS_SPL_stch}.
Initially, the TD3 version is ahead until episode 20, after which the SAC version takes the lead until episode 35.
At the end, TD3-AmelPredSto ends first with 91\% of SPL, followed by SAC-AmelPredSto with 90\%.
DQN-AmelPredSto was unable to surpass a SPL of 10\%.

In the OGN problem, the autonomous UAV must fly around looking for the target object.
The trajectory visualization of each method's flying pattern evolution was used as a qualitative assessment of the navigation problem.
Overall, the DQN algorithm performed the worst at generating a trajectory towards a target.
The best trajectory was achieved by the TD3 algorithm with the AmelPredSto method.
In second place was the SAC-AmelPredSto method since it took more time to perform a more direct path towards the goal pose.
However, more interesting results were obtained in the path obtained by the real-world system to solve the OGN problem described next.

\subsection{Real-world navigation performance}

Real experiment settings were conducted in a flying arena using remote piloted aircraft.
The main focus of real-world experiments was to evaluate the sim2real capability of the proposed method.
The evaluation considered three different target locations with varying altitude levels and positions, focusing on assessing different goal pose aspects such as distance, altitude, and orientation.
The best performing TD3-AmelPredSto model in simulation was used for evaluation in the real flying arena.
Three runs were performed for a single agent, given the increased time required for its execution.
The main bottleneck in the real-world setting was the time required for equipment calibration to acquire accurate positional information, in addition to battery life.
Those aspects were not present in the simulated scenario.

\begin{figure}
    \centering
    \subcaptionbox{}{\includegraphics[width=0.7\columnwidth]{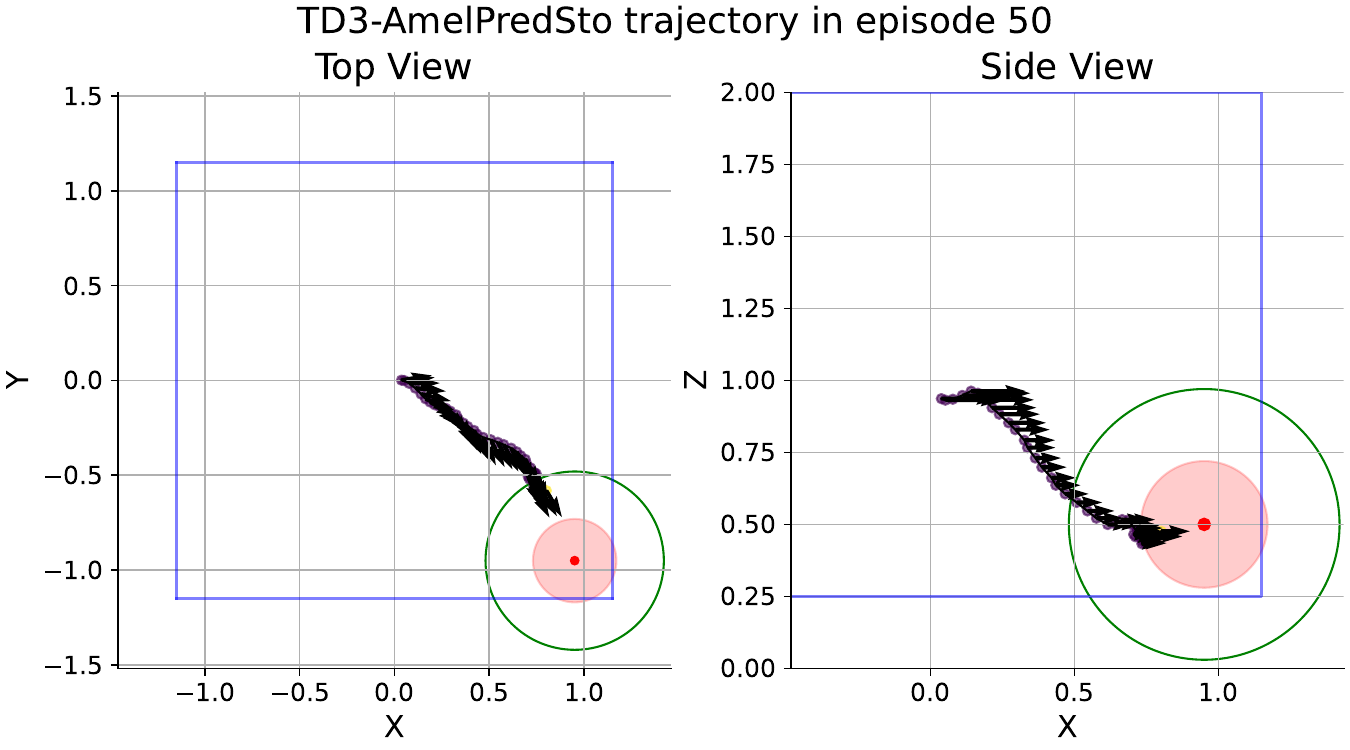}}%

    \subcaptionbox{}{\includegraphics[width=0.7\columnwidth]{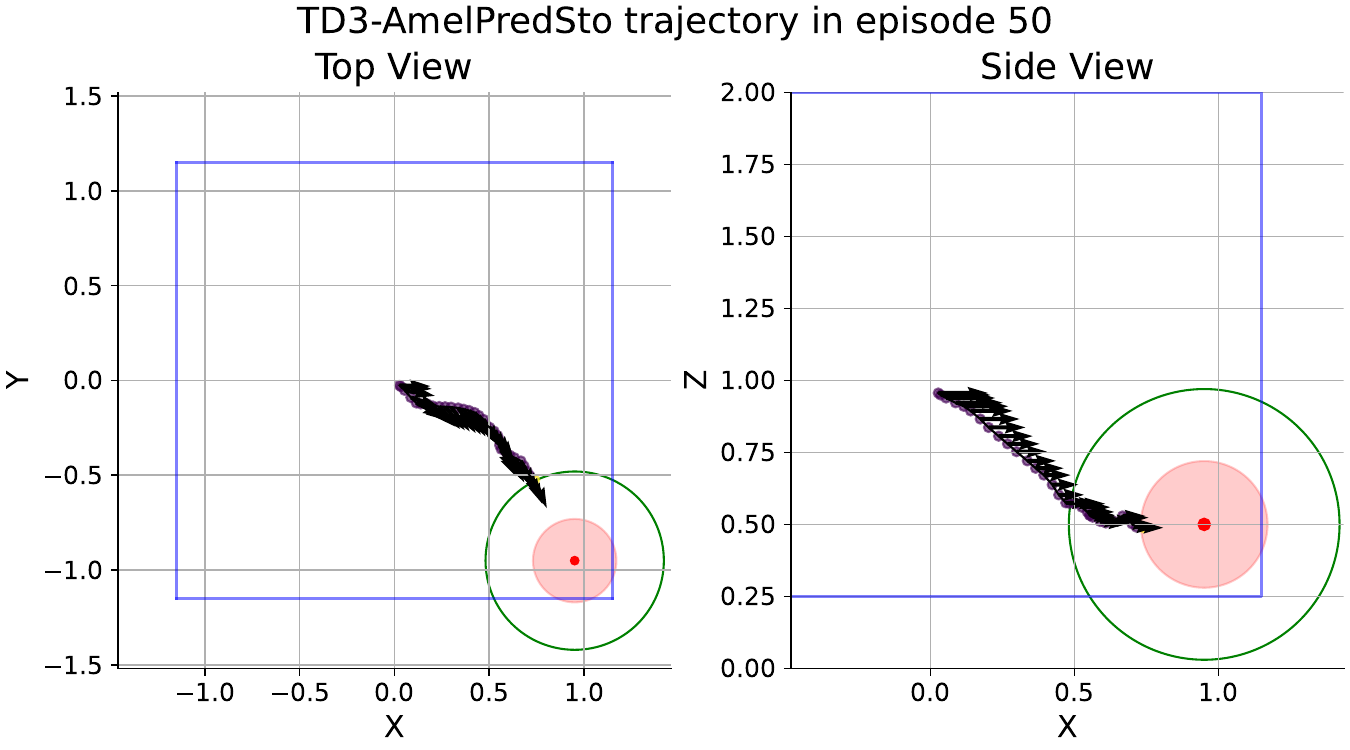}}%

    \subcaptionbox{}{\includegraphics[width=0.7\columnwidth]{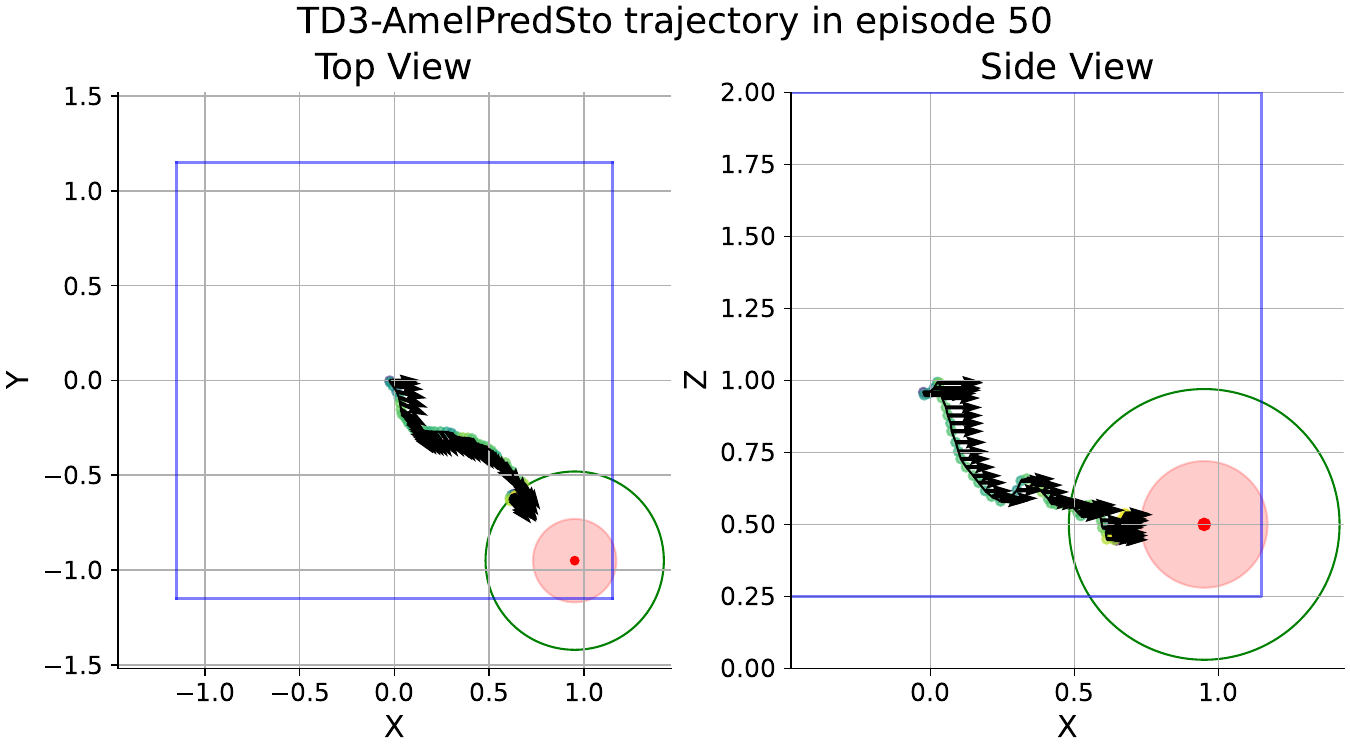}}%
    
    \caption[Real-world trajectories towards the target located at $(0.95, -0.95, 0.5)$ for TD3-AmelPredSto.]{Real-world trajectories towards the target located at $(0.95, -0.95, 0.5)$ for TD3-AmelPredSto.
    Three different runs are depicted in (a), (b), and (c).}\label{fig:path_real_world_01}

\end{figure}

Figure~\ref{fig:path_real_world_01} depicts the performed trajectory of the real Crazyflie in the flying arena towards the target location $(0.95, -0.95, 0.5)$, for $(X, Y, Z)$ coordinates.
A clear path selection can be observed over multiple time steps, in both visible axes pairs $(X, Y)$ and $(X, Z)$ for top and side views, respectively.
The goal zone is defined as the ring-shaped region between the outer green circle and the inner pink-shaded disk, which represents the risk zone.
During the first two attempts, the obtained trajectory pointed directly towards the target.
In contrast, the final attempt struggled to find the correct distance value and expended all the evaluation time.
In this regard, a success rate (SR) of $66.66\%$ was obtained.

\begin{figure}
    \centering
    \subcaptionbox{}{\includegraphics[width=0.7\columnwidth]{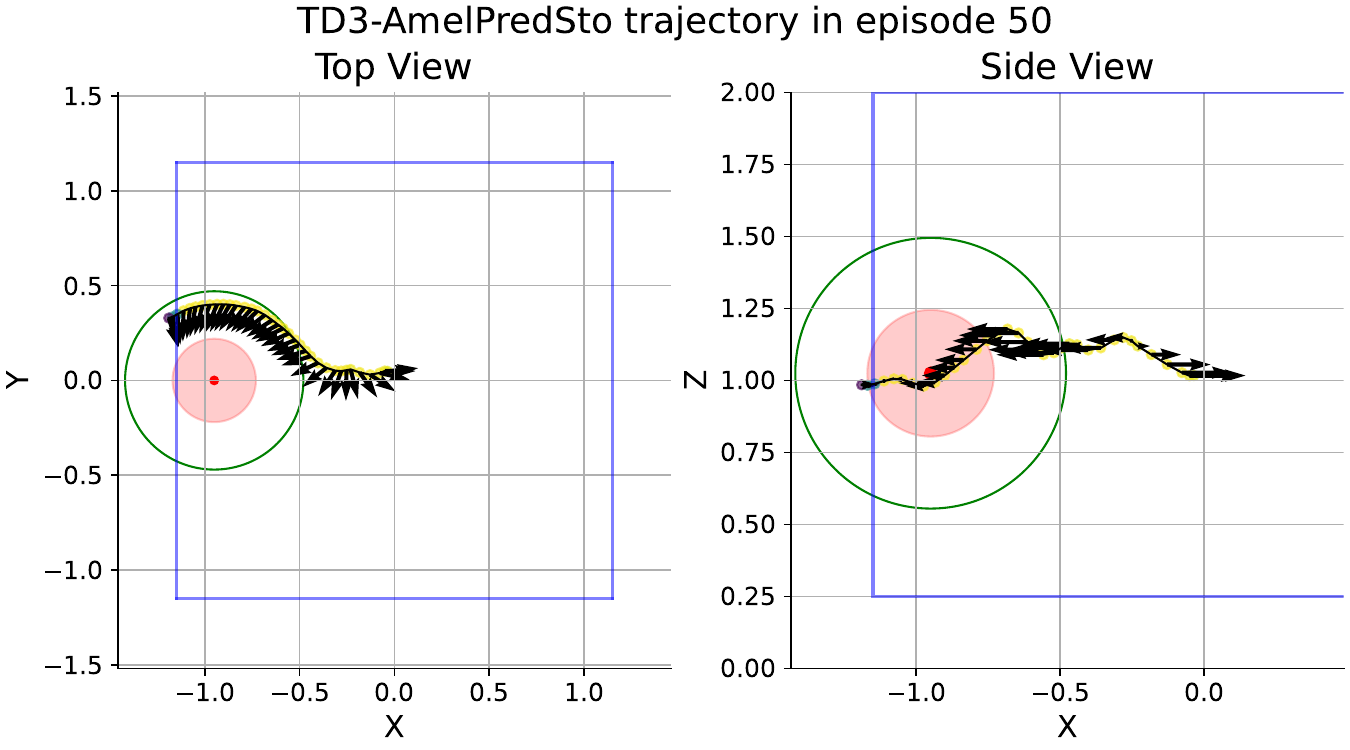}}%

    \subcaptionbox{}{\includegraphics[width=0.7\columnwidth]{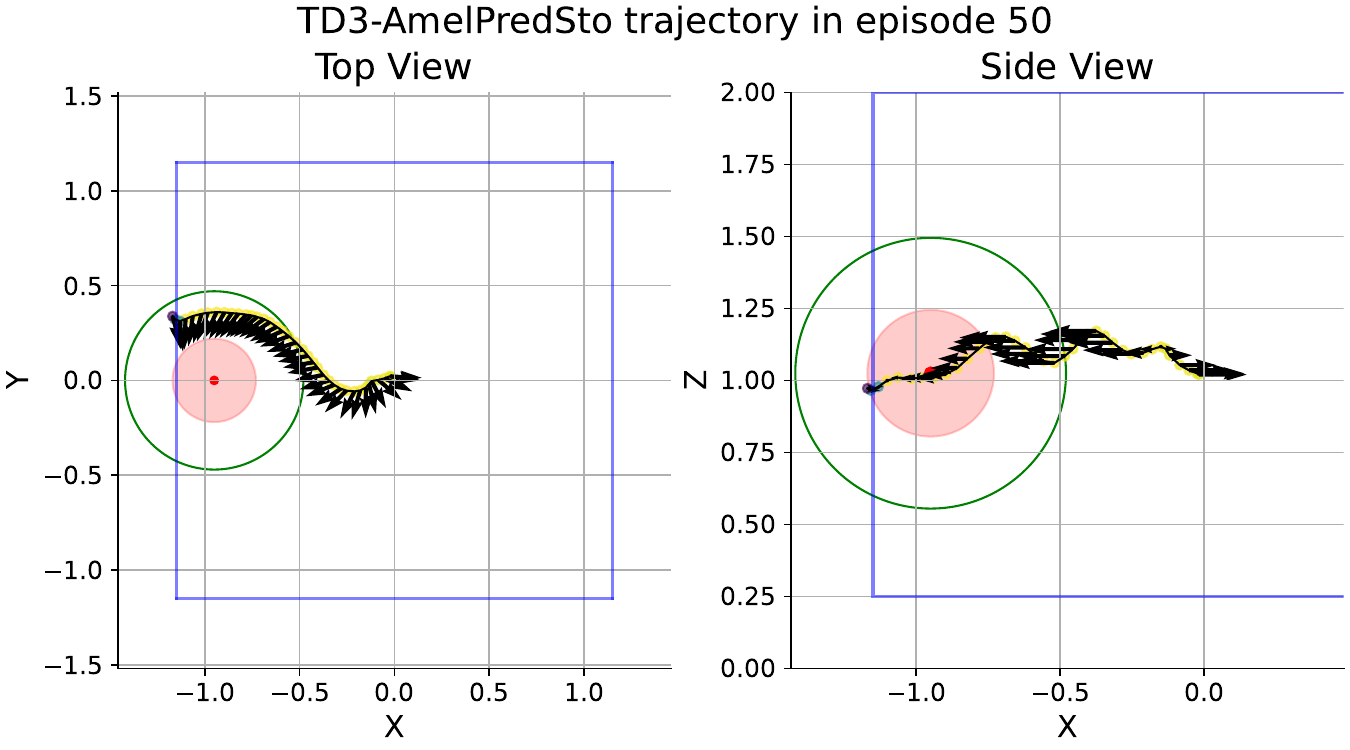}}%

    \subcaptionbox{}{\includegraphics[width=0.7\columnwidth]{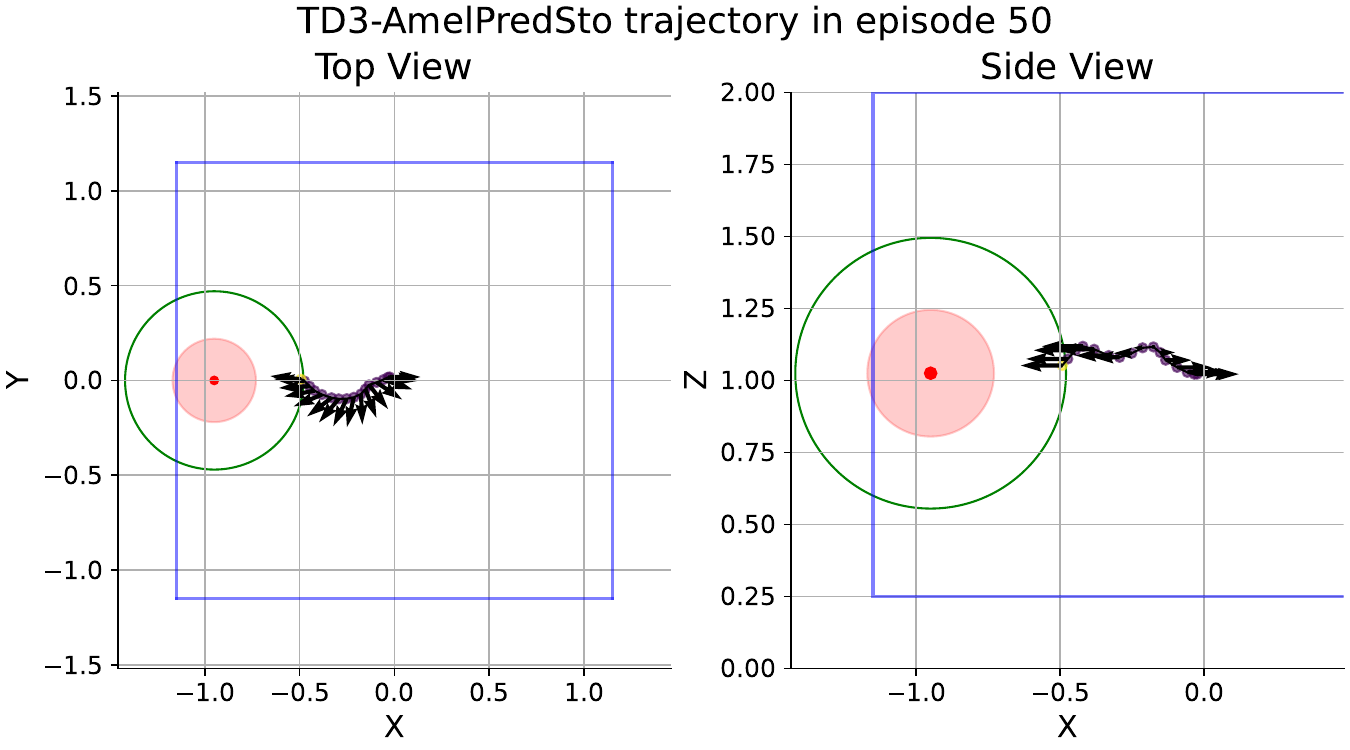}}%
    
    \caption[Real-world trajectories towards the target located at $(-0.95, 0.0, 1.02)$ for TD3-AmelPredSto.]{Real-world trajectories towards the target located at $(-0.95, 0.0, 1.02)$ for TD3-AmelPredSto.
    Three different runs are depicted in (a), (b), and (c).}\label{fig:path_real_world_02}

\end{figure}

Similarly, Figure~\ref{fig:path_real_world_02} depicts the trajectories of the Crazyflie flying towards the target location $(-0.95, 0.0, 1.02)$, for $(X, Y, Z)$ coordinates.
The latest run successfully achieved the goal pose.
All three attempts performed a very similar action sequence from the initial position, diverging near the goal zone.
The first two were capable of keeping the vehicle within the goal zone and orbiting the target location.
However, they exceeded the flight area represented by the blue box.
The last one was the only one capable of entering with a more correct orientation.
Such behavior can result from the reduced motion's complexity, since the altitude component does not need to be adjusted.

\begin{figure}
    \centering
    \subcaptionbox{}{\includegraphics[width=0.7\columnwidth]{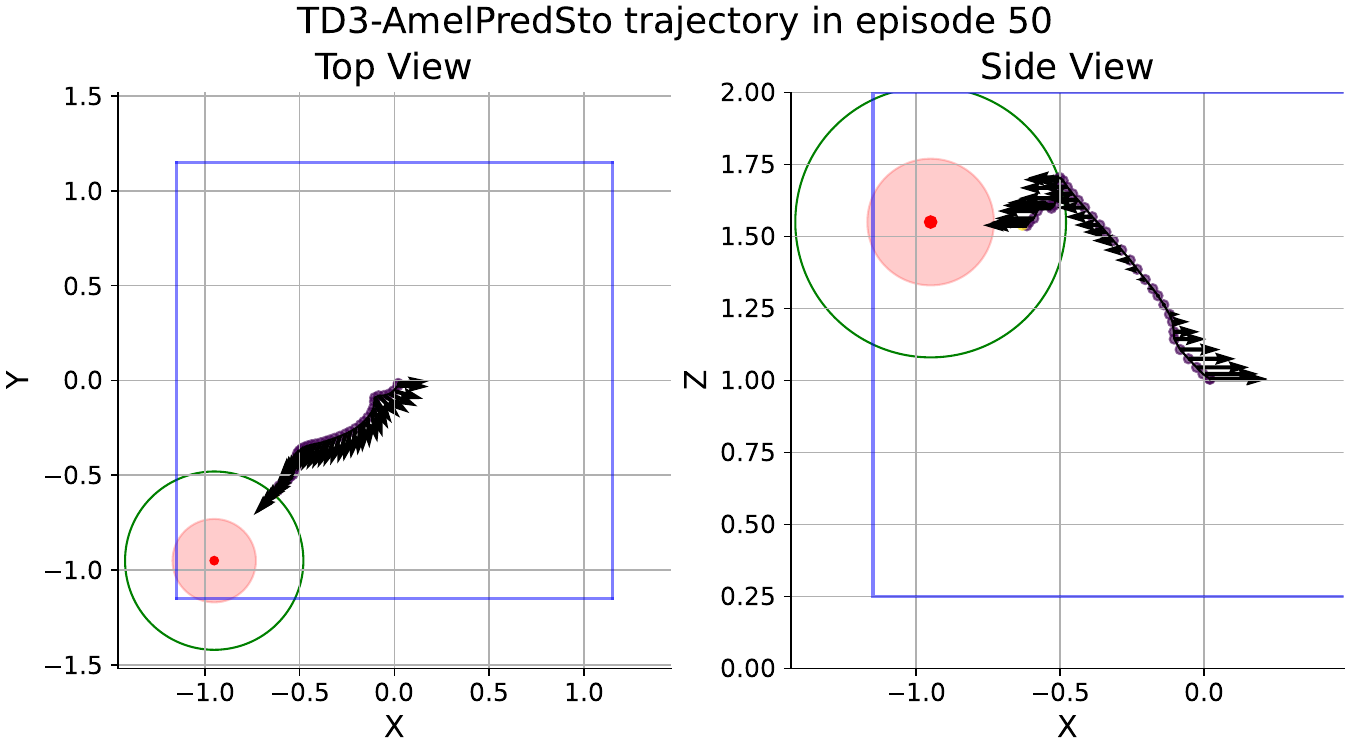}}%

    \subcaptionbox{}{\includegraphics[width=0.7\columnwidth]{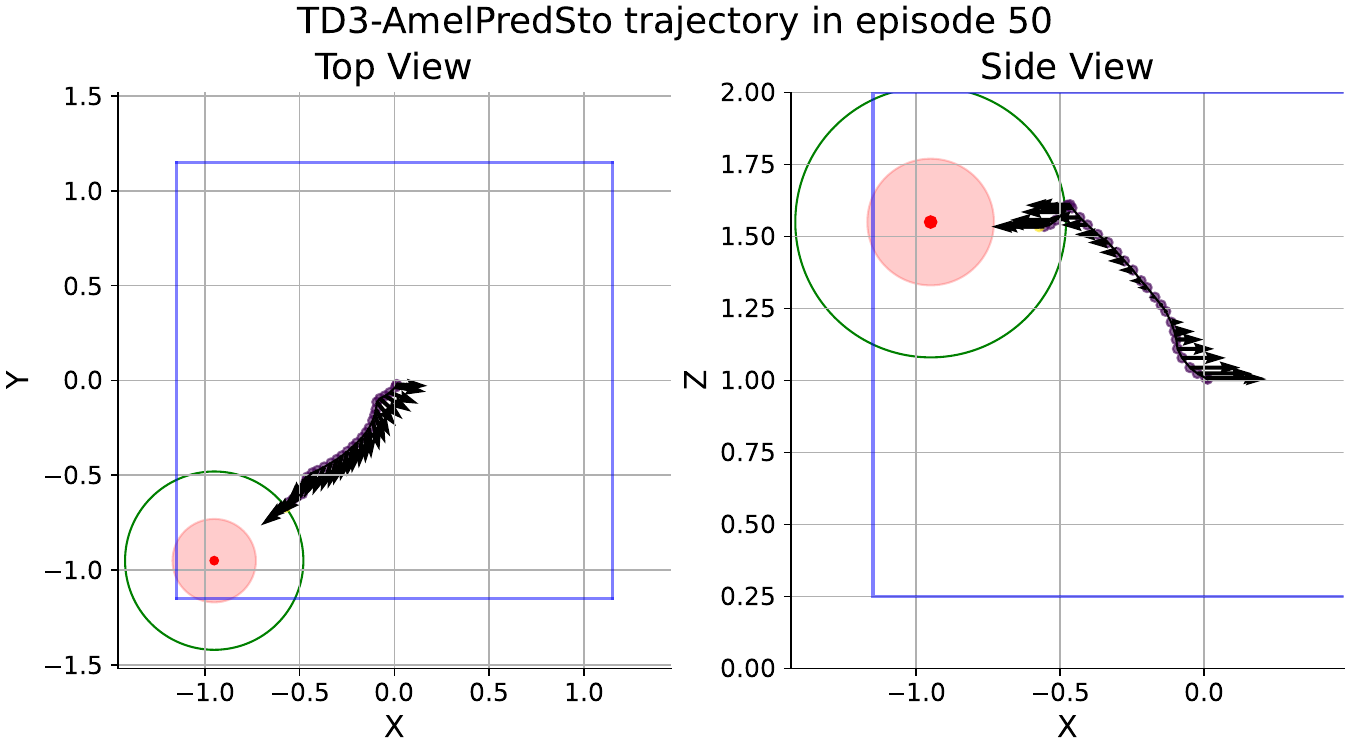}}%

    \subcaptionbox{}{\includegraphics[width=0.7\columnwidth]{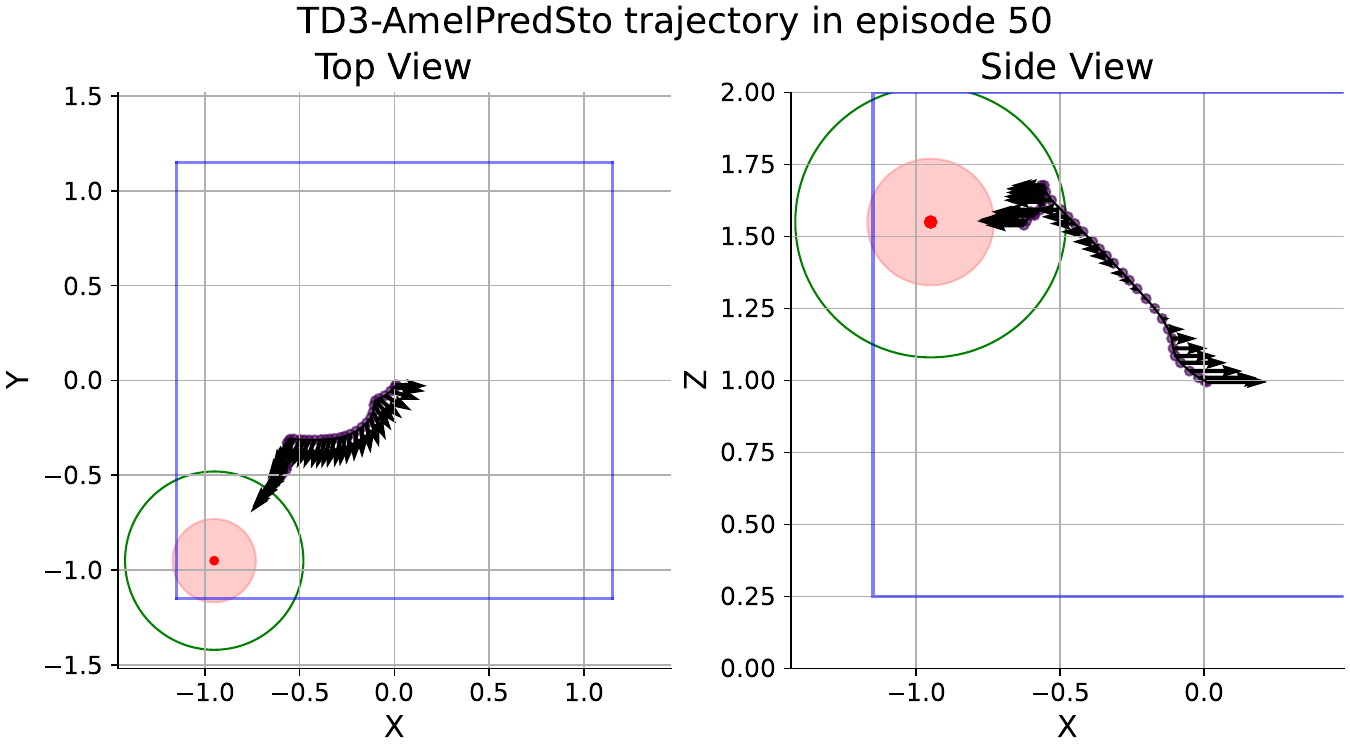}}%
    
    \caption[Real-world trajectories towards the target located at $(-0.95, -0.95, 1.55)$ for TD3-AmelPredSto.]{Real-world trajectories towards the target located at $(-0.95, -0.95, 1.55)$ for TD3-AmelPredSto.
    Three different runs are depicted in (a), (b), and (c).}\label{fig:path_real_world_03}

\end{figure}

Finally, Figure~\ref{fig:path_real_world_03} depicts the trajectories of the Crazyflie flying towards the target location $(-0.95, -0.95, 1.55)$, for $(X, Y, Z)$ coordinates.
In all three instances, the vehicle corrected its altitude and orientation towards the target until it surpassed it and approached the desired pose.
In general comparison, correcting the orientation is simpler in cases where the traveled distance is also high.
Both target locations $(-0.95, -0.95, 1.55)$ and $(0.95, -0.95, 0.5)$ were diagonally opposed.
Hence, the traveled distance is higher, and TD3-AmelPredSto performed the best.
In contrast, for the target location $(-0.95, 0.0, 1.02)$, where it must practically change its orientation the most when traveling, TD3-AmelPredSto performed the worst.

Overall real-world results obtained, demonstrated a suitable sim2real transfer capability without with TD3-AmelPredSto performing a SPL of $65.90\%$, and a DTS of $0.10$ meters.
Therefore, no adjustments or fine-tunning of trained model in simulated environment were required for such real-world tests, achieving a suitable generalization capabilities.
Additionally, the use of the Crazyflie's firmware for low-level control also ensured the sim2real transfer compatibility.

\section{Conclusions}~\label{section:conclusion}
Autonomous UAV navigation under an object-goal context is challenging because the flight pattern must ensure vehicle displacement to reach a desired object.
SRL aids learning by decoupling representation and policy learning from raw input data, enhancing the sample efficient.
In terms of OGN performance, it was observed that continuous actor-critic RL algorithms were the best.
Overall, the DQN outcome was not satisfactory, despite the higher reward and efficiency achieved by DQN-AmelPredDet, which surpassed the SPL by 80\% compared to its vanilla version.
We hypothesize that the reason could be a low DTS achieved between the UAV and the target, given the discrete action signal.
The AmelPredSto method was the most sample-efficient, constantly enhancing the number of times each agent reached the goal pose.
A positive change was observed for SAC-AmelPredSto, which significantly reduces the standard deviation in metrics related to navigation.
Nevertheless, the AmelPredDet method ultimately diminishes its performance compared to the AmelPredSto methods.
Only TD3-AmelPredDet was able to surpass its stochastic version in SPL at the end, possibly due to better compatibility with the policy function.

TD3-AmelPredSto achieved a remarkably successful outcome.
From the very beginning, it was able to identify and slowly displace the UAV towards the goal pose.
A prioritized adjustment in orientation and altitude components of the goal pose can interpret the evolution to a far distance.
Ultimately, it can reach a more distant and safe position with a more direct orientation related to the difference between the target's location and the initial position.
This safest behavior is responsible for diminishing the SPL of TD3-AmelPredSto, as proved by the DTS, which prioritizes larger distance values as quickly as possible.
So far, TD3-AmelPredSto is the most suitable option for a sample-efficient object-goal autonomous navigation system.
Experiments in real-world settings demonstrated a suitable sim2real transferability capable of reaching the goal pose with an SR of $66.66\%$, a SPL of $65.90\%$, and a DTS of $0.10$ meters.

Future work should be addressed in large-scale dynamic scenarios including the presence of wind, obstacles, and moving targets to assess more complex navigation settings.
Other studies can focus on multi-agent settings where the latent representation must also be aware of other agents in a cooperative or competitive cases.
Additionally, the AmelPredSto assessment must be carried out in other benchmark problems such as MuJoCo and Atari.

\section*{Acknowledgment}
This study was financed in part by the Coordenação de Aperfeiçoamento de Pessoal de Nível Superior - Brasil (CAPES) - Finance Code 001,
Fundação de Amparo a Ciência e Tecnologia do Estado de Pernambuco (FACEPE),
and Conselho Nacional de Desenvolvimento Científico e Tecnológico (CNPq) - Brazilian research agencies.

\bibliographystyle{IEEEtran}
\bibliography{references}

\end{document}